\documentclass[10pt,twocolumn,letterpaper]{article}

\usepackage{iccv}              % To produce the CAMERA-READY version
\usepackage{subcaption} % 複数の図を並べる
\usepackage{float} % 図表配置のHオプションを指定可能にする
\usepackage{booktabs} % 表の横罫線
\usepackage{multirow} % 複数行をまとめる
\usepackage{makecell} % セル内での改行
\usepackage{xcolor}
\usepackage{tikz}
\usepackage{tabularx}
\usepackage{tcolorbox}

\usepackage[disable]{todonotes} % TODOを無効にする = 本提出時

\usepackage[inline]{enumitem} % 箇条書きの調整

\usepackage{amsmath,amssymb}

\usepackage{xspace}
\usepackage{url}
\newcommand{\term}[1]{\ifmmode \mathit{#1}\xspace\else\textit{#1}\xspace\fi}

\def\TOPIC#1{} % トピックを無効にする場合 = 本提出時

\newcommand{\egooops}{EgoOops\xspace}

\newcommand{\electricalcircuit}{EC\xspace}
\newcommand{\colormixture}{CM\xspace}
\newcommand{\ionicreaction}{IR\xspace}
\newcommand{\buildingblock}{BB\xspace}
\newcommand{\cardboard}{CB\xspace}

\newcommand{\ordermistakes}{order mistakes\xspace}
\newcommand{\executionmistakes}{execution mistakes\xspace}

\newcommand{\missing}{Missing\xspace}
\newcommand{\outoforder}{Out-of-order\xspace}
\newcommand{\resume}{Pause-and-resume\xspace}
\newcommand{\undefined}{Undefined-step\xspace}

\newcommand{\object}{Object\xspace}
\newcommand{\mispick}{Mispick\xspace}
\newcommand{\correction}{Correction\xspace}
\newcommand{\accident}{Accident\xspace}
\newcommand{\howto}{Way\xspace}
\newcommand{\others}{Others\xspace}

\definecolor{iccvblue}{rgb}{0.21,0.49,0.74}
\usepackage[pagebackref,breaklinks,colorlinks,allcolors=iccvblue]{hyperref}

\def\paperID{9} % *** Enter the Paper ID here
\def\confName{ICCV}
\def\confYear{2025}

\title{EgoOops: A Dataset for Mistake Action Detection \\from Egocentric Videos referring to Procedural Texts}

\author{
Yuto Haneji${}^{1}$, Taichi Nishimura${}^{2}$, Hirotaka Kameko${}^{1}$, Keisuke Shirai${}^{1}$, Tomoya Yoshida${}^{1}$, \\
Keiya Kajimura${}^{1}$, Koki Yamamoto${}^{1}$, 
Taiyu Cui${}^{1}$, Tomohiro Nishimoto${}^{1}$, Shinsuke Mori${}^{1}$\\
${}^{1}$Kyoto University, ${}^{2}$Sony Interactive Entertainment\\
{\tt\small haneji.yuto.s66@kyoto-u.jp, Taichi.A.Nishimura@sony.com,}\\
{\tt\small shirai.keisuke.5y@kyoto-u.ac.jp, \{kameko,forest\}@i.kyoto-u.ac.jp}\\
}

\begin{document}
% \maketitle
\twocolumn[{
    \renewcommand\twocolumn[1][]{#1}
    \maketitle
    \begin{center}
        \newcommand{\teaserwidth}{\textwidth}
        \centerline{
            \includegraphics[width=\teaserwidth]{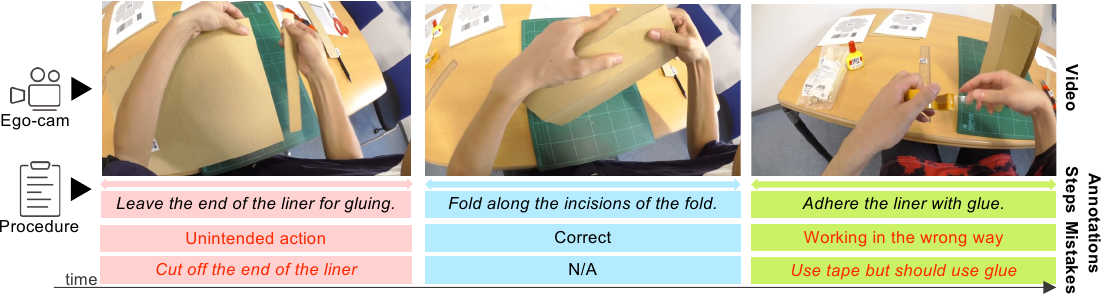}
        }
        \captionsetup{type=figure}
        \caption{An example task (cardboard) from our EgoOops dataset. EgoOops includes 50 egocentric videos across five procedural domains and corresponding procedural texts. It contains three types of annotations: video-text alignment, mistake labels, and descriptions explaining the errors in each segment.}
        \label{fig:teaser}
    \end{center}
}]
 % \maketitle is done inside here.
\begin{abstract}
Mistake action detection is crucial for developing intelligent archives that detect workers' errors and provide feedback. Existing studies have focused on visually apparent mistakes in free-style activities, resulting in video-only approaches to mistake detection. However, in text-following activities, models cannot determine the correctness of some actions without referring to the texts. Additionally, current mistake datasets rarely use procedural texts for video recording except for cooking. To fill these gaps, this paper proposes the EgoOops dataset, where egocentric videos record erroneous activities when following procedural texts across diverse domains. It features three types of annotations: video-text alignment, mistake labels, and descriptions for mistakes. We also propose a mistake detection approach, combining video-text alignment and mistake label classification to leverage the texts. Our experimental results show that incorporating procedural texts is essential for mistake detection. Data is available through \url{https://y-haneji.github.io/EgoOops-project-page/}.
\end{abstract}
\section{Introduction}
\label{sec:intro}

\newcolumntype{C}{>{\centering\arraybackslash}p{5.5em}}
\begin{table*}[t]
    \centering
    \caption{Comparison of mistake datasets. 
    % EgoOops is the first dataset of egocentric videos annotated with order and execution mistakes in following procedural texts across diverse domains.
    % EgoOops is the first dataset where egocentric videos following procedural texts across diverse domains are annotated with categories and descriptions of order and execution mistakes. 
    Range: mistake labels to a video or each step segment with start and end times; Cat.: finer-grained categorization than \term{correct}, \term{mistake}, and \term{correction}; Desc.: descriptions explaining why each segment is incorrect; OM, EM: order mistakes, execution mistakes (see \cref{subsec:term}); Proc.: workers follow step-by-step procedural texts. $^{\ast}$Domain specific (\eg, extra screws). $^{\dagger}$Orders unseen in the training set, not always faulty.}
    \label{tab:comp-rel-data}
    \begin{tabular}{l|ccccc|cccrr} \toprule
        \multirow{2}{*}{Dataset} & \multicolumn{5}{c|}{Mistake annotations} & \multirow{2}{*}{Domain} & \multirow{2}{*}{Proc.} & \multirow{2}{*}{Ego} & \multirow{2}{*}{\#videos} & \multirow{2}{*}{Duration (hour)} \\
         & Range & Cat. & Desc. & OM & EM & & & & & \\
        \midrule
        % GTEA & kitchen & $\times$ & $\checkmark$ & $\checkmark$ & $\checkmark$ & 2012\\
        % 50Salads & kitchen & $\times$ & $\times$ & $\checkmark$ & $\checkmark$ & 2013\\
        % What's Cooking & kitchen & $\checkmark$ & $\times$ & $\times$ & $\times$ & $\times$ & 2015 & \\
        % Charades & daily activities & $\times$ & $\times$ & $\checkmark$ & $\checkmark$ & 2016\\
        % EGTEA & kitchen & $\times$ & $\checkmark$ & $\checkmark$ & $\checkmark$ & 2018\\
        % YouCook2~\cite{zhou_towards_2018} &  \multicolumn{5}{c|}{\multirow{5}{*}{$\times$}} &Cooking& $\checkmark$ & $\times$ & 2,000 & 176\phantom{.0} \\
        % CrossTask~\cite{zhukov_cross-task_2019} & & & & & &Diverse& $\times$ & $\times$ & 4,700 & 375\phantom{.0} \\
        % HowTo100M~\cite{miech_howto100m_2019} & & & & & & Diverse& $\checkmark$ & $\times$ & 1.221M & 134,472\phantom{.0} \\
        % EPIC-KITCHENS~\cite{damen_epic-kitchens_2021} & & & & & &Cooking& $\checkmark$ & $\checkmark$ & 432 & 55\phantom{.0}\\
        % BioVL~\cite{nishimura_egocentric_2021} & & & & & & Biochemistry & $\checkmark$ & $\checkmark$ & 16 & 1.6\\
        % % Ego4D & daily activities & $\times$ & $\checkmark$ & $\times$ & $\checkmark$ & 2022\\
        % \midrule
        Assembly101~\cite{sener_assembly101_2022} & Segment & $\times$ & $\times$ & $\checkmark$ & $\times$ & Assembly& $\times$ & $\checkmark$ & 1,425 & 167\phantom{.0} \\ 
        ATA~\cite{ghoddoosian_weakly-supervised_2023} & Video & $\checkmark^{\ast}$ & $\times$ & $\checkmark^{\dagger}$ & $\checkmark$ & Assembly & $\times$ & $\times$ & 1,152 & 24.8\\
        HoloAssist~\cite{wang_holoassist_2023} & Segment & $\times$ & $\checkmark$ & $\times$ & $\checkmark$ & Assembly & $\times$ & $\checkmark$ & 2,221 & 166\phantom{.0} \\
        IndustReal~\cite{schoonbeek_industreal_2024} & Segment & $\times$ & $\checkmark$ & $\checkmark$ & $\checkmark$ & Assembly & $\times$ & $\checkmark$ & 84 & 5.8 \\
        EgoPER~\cite{lee_error_2024} & Segment & $\checkmark$ & $\checkmark$ & $\checkmark$ & $\checkmark$ & Cooking & $\checkmark$ & $\checkmark$ & 386 & 28\phantom{.0} \\
        CaptainCook4D~\cite{peddi_captaincook4d_2024} & Segment & $\checkmark^{\ast}$ & $\checkmark$ & $\checkmark$ & $\checkmark$ & Cooking & $\checkmark$ & $\checkmark$ & 384 & 94.5\\
        \midrule
        \textbf{EgoOops (Ours)} &  Segment & $\checkmark$ & $\checkmark$ & $\checkmark$ & $\checkmark$ & Diverse& $\checkmark$ & $\checkmark$ & 50 & 6.8\\ \bottomrule
    \end{tabular}
\end{table*}

% Topic: general background
% When performing procedural tasks such as assembly or scientific experiments, we follow procedural texts to carry them out in the real world. 
Procedural activities are common in daily life and expert fields, such as assembly, experimentation, and cooking. People often carry them out by following procedural texts in the real world. During this process, mistakes negatively impact quality, speed, cost, and safety. Common mistakes include skipped necessary steps or wrong execution ways, which can sometimes result in life-or-death situations. One promising solution to this problem is to develop an intelligent video archive that records workers' activities, detects their mistakes, and shows them the mistake clips to prevent recurrences.

% Topic: related work / research map 手続きタスクの撮影と映像データセットの構築
The intelligent video archives are developed using video datasets recording workers' steps in detail. Many egocentric video datasets~\cite{damen_epic-kitchens_2021,nishimura_egocentric_2021,bansal_my_2022,grauman_ego4d_2022,sener_assembly101_2022,ragusa_meccano_2023, wang_holoassist_2023,peddi_captaincook4d_2024,schoonbeek_industreal_2024,lee_error_2024} % sigurdsson_actor_2018,li_eye_2018
have been proposed by equipping a worker with an egocentric camera to capture activity details. Previously, most datasets were interested in correct execution of activities~\cite{damen_epic-kitchens_2021,nishimura_egocentric_2021,grauman_ego4d_2022,bansal_my_2022,ragusa_meccano_2023}. % sigurdsson_actor_2018,li_eye_2018,
Some recent studies have also included and annotated mistake actions in their assembly~\cite{sener_assembly101_2022,wang_holoassist_2023,ghoddoosian_weakly-supervised_2023,schoonbeek_industreal_2024} and cooking~\cite{peddi_captaincook4d_2024,lee_error_2024} video datasets. Using these mistake video datasets, researchers have proposed mistake detection approaches~\cite{sener_assembly101_2022,ghoddoosian_weakly-supervised_2023,peddi_captaincook4d_2024,flaborea_prego_2024,seminara_differentiable_2024-1,lee_error_2024}. % ding_every_2023

% Topic: issues
However, these studies have the following three limitations (see \cref{tab:comp-rel-data}). 
\begin{enumerate*}[label=\textbf{L\arabic*:},%align=left,nosep,labelwidth=1em,itemindent=1em+\labelsep,leftmargin=0pt
]
    \item \textbf{video-focused approaches.} Existing studies~\cite{sener_assembly101_2022,ghoddoosian_weakly-supervised_2023,peddi_captaincook4d_2024,flaborea_prego_2024,seminara_differentiable_2024-1,lee_error_2024,huang_modeling_2025} % ding_every_2023 
    have focused on video-only approaches to mistake detection and not utilized procedural texts. Their approaches are suitable for visually apparent mistakes like incorrectly attached parts and falling plates. Nevertheless, in text-following activities, some mistakes are deviations from procedural texts, thus not obvious only from visual cues (\eg, in \cref{fig:teaser}, the use of tape is a mistake because the text designates glue). Therefore, besides videos, texts are essential for models to detect mistakes accurately.
    \item \textbf{specific domains.} Procedural texts are included in only a few mistake datasets recording cooking~\cite{peddi_captaincook4d_2024,lee_error_2024}. Since many real-world activities follow procedural texts, it is essential to collect data from more domains.
    \item \textbf{rough mistake labels.} Most datasets define coarse-grained (\term{correct}/\term{mistake}/\term{correction})~\cite{sener_assembly101_2022,wang_holoassist_2023,schoonbeek_industreal_2024} or domain-specific (\eg, \term{extra screws})~\cite{ghoddoosian_weakly-supervised_2023,peddi_captaincook4d_2024} categories. General and fine-grained categorization enables analysis of mistake patterns across diverse domains (\eg, a commonly frequent category of mistakes). 
\end{enumerate*}

% Topic: this paper (dataset)
To address these issues, we propose a novel dataset called \textbf{EgoOops} (see \cref{fig:teaser}), where egocentric videos record erroneous activities when following procedural texts (L1) across diverse domains (L2). 
Given the collected videos and texts, we perform the following three steps for annotations. 
First, we align steps in the procedural text with video segments (\ie, start and end timestamps). Second, if a segment contains a mistake action, we categorize it into six mistake classes (L3). Finally, the segments assigned mistake labels are provided with descriptions of why the actions are considered mistakes.
As for the size and tasks, EgoOops contains 50 egocentric videos totaling 6.8 hours across five tasks of new domains: electrical circuits, color mixture experiments, ionic reaction experiments, toy building blocks, and cardboard crafts.

% Topic: this paper (models, experiments)
We also propose an approach to the problem of mistake action detection, especially focusing on the utilization of procedural texts (L1). To leverage the texts, our approach combines video-text alignment and mistake label classification; the former localizes the start and end times of each procedural step, and the latter assigns the step segments one label of mistake classes. 
In experiments using EgoOops, our multi-modal approach outperforms a video-only baseline, and the ablation of textual inputs decreases our performance. These results demonstrate that incorporating procedural texts is essential for mistake detection.
Additionally, we test existing mistake classifiers and multi-modal large language models for the classification problem. As for the alignment problem, we compare our fully-supervised approach with zero-shot and self-supervised ones. The results confirm that EgoOops targets novel domains of procedural activities and contains useful alignment annotations.

\section{Related work}

In this section, we compare EgoOops with other datasets in terms of two perspectives: procedural activity dataset and mistake action dataset. % \Cref{tab:comp-rel-data} shows comparisons of EgoOops and other datasets.

\subsection{Procedural activity datasets}
Procedural activity understanding is an important capability for enhancing smart systems, such as AR/VR assistants~\cite{wang_holoassist_2023} and intelligent archives. In particular, aligning a sequence of step instructions with a video (\ie, video-text alignment) is fundamental.
To support research in this direction, a variety of procedural activity datasets have been developed. While early datasets collected third-person perspective videos accompanied by textual descriptions of each timestep from YouTube~\cite{zhukov_cross-task_2019,tang_coin_2019,miech_howto100m_2019}, recent datasets focus on first-person (egocentric) videos that capture fine-grained details of workers' activities~\cite{sigurdsson_actor_2018,li_eye_2018,damen_epic-kitchens_2021,nishimura_egocentric_2021,ragusa_meccano_2023,grauman_ego4d_2022,bansal_my_2022}.
For example, EPIC-KITCHENS~\cite{damen_epic-kitchens_2021} dataset consists of 432 egocentric videos with step instructions in the cooking domain, including start and end times for each segment. 
Our dataset provides similar video-text alignment annotations as existing datasets, but distinguishes itself by focusing on erroneous actions, thereby enabling the study of error detection within procedural activities.

\subsection{Mistake action datasets}
Mistakes are critical in procedural activities, as they can propagate through subsequent steps and significantly affect task success. This has led to the development of various datasets (see \cref{tab:comp-rel-data}). These datasets are categorized into free-style and text-following settings.

% Free-style
In free-style activity datasets~\cite{sener_assembly101_2022,ghoddoosian_weakly-supervised_2023,wang_holoassist_2023,schoonbeek_industreal_2024}, workers aim to complete goals not relying on procedural texts. %, and procedural texts do not strictly define the steps or not even exist. 
For example, Assembly101~\cite{sener_assembly101_2022} records toy assembly and annotates segment-level labels of \term{correct}, \term{mistake}, and \term{correction}.
% HoloAssist~\cite{wang_holoassist_2023} records object manipulation, where instructors verbally intervene in mistakes through a mixed-reality headset.
Building on such datasets, previous studies have proposed various approaches to find mistakes in videos~\cite{
sener_assembly101_2022,%
% ding_every_2023,%
ghoddoosian_weakly-supervised_2023,%
wang_holoassist_2023,%
flaborea_prego_2024,%
seminara_differentiable_2024-1%
}. These approaches rely only on videos because mistakes in free-style activities are mainly visually apparent such as incorrectly attached parts and dropped screws.

% Procedural texts (ours)
Another line of studies records workers following procedural texts in their datasets~\cite{peddi_captaincook4d_2024,lee_error_2024}. EgoPER~\cite{lee_error_2024} annotates recipe-execution videos with order and execution mistakes, and CaptainCook4D~\cite{peddi_captaincook4d_2024} proposes categorization specific to cooking (\eg, \term{temperature error}).
One problem is that they still do not utilize procedural texts to find mistakes in videos. In text-following activities, some mistakes are deviations from procedural texts, thus the texts are essential for models. In addition, to the best of our knowledge, no datasets besides cooking involve procedural texts, whereas many real-world activities follow them.
Our EgoOops dataset annotates video-text alignment and mistake labels across diverse domains, promoting the utilization of texts in mistake detection.

%\section{EgoOops dataset}
\section{\egooops dataset}

\egooops dataset provides procedural activity videos including erroneous work and annotations for these mistakes. The activities are performed by following instructional steps of procedural texts in order. In this section, we first define mistakes and then describe task selection, video recording, and annotations. Finally, dataset statistics are provided.

%\subsection{Mistake definition}
%\label{subsec:term}
%We define \term{mistakes} as deviations from a procedural text. This definition leads to two mistake types: \term{\ordermistakes} and \term{\executionmistakes}. 
%Order mistakes occur when the steps executed by a worker are different from the ones listed in the procedural text. These include skipping necessary steps (\term{missing}), swapping the step order (\term{out-of-order}), inserting extra steps (\term{undefined}), and pausing a step and resuming after others (\term{split}).
%Execution mistakes occur when a worker fails to follow the instructions while carrying out a step. We categorize them into the following six classes: 
%\begin{enumerate*}[label=(\arabic*)]
%    \item working with wrong objects (\term{\object}), 
%    \item grasping wrong objects and releasing them without using (\term{\mispick}), 
%    \item correction of mistake actions (\term{\correction}), 
%    \item unintended actions (\term{\accident}), 
%    \item performing in the wrong way (\term{\howto}), 
%    \item others (\term{\others}). 
%\end{enumerate*}

%%%%%%%%%%%%%%%%%%%%%
% 修論の記述
%%%%%%%%%%%%%%%%%%%%%

\subsection{Mistake definition}
\label{subsec:term}

%We define activity mistake as ``deviation from instructional steps.'' In this study, we focus on the two types of the activity mistakes: \textbf{execution mistake} and \textbf{order mistake}.
We define mistakes as ``deviations from instructional steps.'' Out of mistake types that meet this definition, we consider executions mistakes in this study.\footnote{Other types of mistakes such as order mistakes (e.g., skipping and reordering steps) may also occur during activities, but we focus on the execution mistakes which are our main concern. We leave the study of the other types of mistakes to future work.}

Execution mistakes occur when a worker misinterprets and executes steps. We classify the execution mistakes into the six types of errors:
\begin{enumerate}[nosep]
    \item \textbf{Incorrect-object-use (\term{\object})} executes a step with a different object specified in the text. This includes cases when the incorrect number of objects is used.
    \item \textbf{Incorrect-object-picking (\term{\mispick})} picks up an incorrect object, but the worker recognizes the mistake and releases the object. This mistake does not execute a step, unlike incorrect-object-use.
    % \item \textbf{Self-correction} (\term{correction}) executes a step in a wrong way and recognizes and corrects the mistake.
    \item \textbf{Self-correction (\term{\correction})} recognizes and corrects mistakes in a step that has been executed in a wrong way.
    \item \textbf{Accidental-mistakes (\term{\accident})} causes accidental happenings mainly due to carelessness.
    \item \textbf{Wrong-way (\term{\howto})} picks a correct object but executes a step in a way that misaligns with the instruction in the text.
    \item \textbf{Other-mistakes (\term{\others})} induces other types of mistakes. This also includes cases where multiple types of the above mistakes occur simultaneously.
\end{enumerate}

% なお，本研究で目指すのは，\executionmistakes を手順書を参照しながら検出することである．しかし，\ordermistakes も，EgoOopsデータセットの映像には含まれている．これらのアノテーションも付与し，後続の研究において活用されることを期待する．

%Order mistake happens when a worker executes steps in a different order than written in procedural texts. We classify the order mistakes into three cases: skipping steps, reordering steps, interrupt-then-resume steps.\footnote{In our preliminary experiment, we found cases that a worker interrupts a step midway, completes other steps, and then resume the interrupted step.}

%%%%%%%%%%%%%%%%%%%%%

\subsection{Task selection}
\label{subsec:task-selection}

% タスク選定
%To record procedural activities in diverse domains with various mistake actions, we selected the following five tasks: electrical circuits (\term{\electricalcircuit}), color mixture experiments (\term{\colormixture}), ionic reaction experiments (\term{\ionicreaction}), toy building blocks (\term{\buildingblock}), and cardboard crafts (\term{\cardboard}). Their domains are diverse (\ie, electronics, chemistry, assembly, and crafts), and various mistakes can happen (\eg, using wrong chemicals, accidentally cutting off cardboard).

To capture a broad range of mistake types described in \cref{subsec:term}, we construct the \egooops dataset with activity videos from diverse domains. Accordingly, we select the following five tasks:
 \begin{itemize}[nosep]
     \item \textbf{Electrical circuits} (\textbf{\term{\electricalcircuit}}): Connecting electrical elements to complete an electrical circuit that turns a propeller.
     \item \textbf{Color mixture experiments} (\textbf{\term{\colormixture}}): Examining the color of various solutions of detergent and fluorescent paint when illuminating them with a blacklight. 
     \item \textbf{Ionic reaction experiments} (\textbf{\term{\ionicreaction}}): Examining ionic reaction by dropping chemical solutions to metal plates.
     \item \textbf{Toy building blocks} (\textbf{\term{\buildingblock}}): Piling up building blocks to construct the specified structure.
     \item \textbf{Cardboard crafts} (\textbf{\term{\cardboard}}): Crafting Omikuji boxes, Japanese random fortunes, from cardboard.
 \end{itemize}

We prepare procedural texts for each task as follows:  
\colormixture and \cardboard use procedures collected from the web;  
\ionicreaction and \electricalcircuit use instruction manuals included with out-of-box kits;  
\buildingblock uses a procedure that we manually write from scratch.

\subsection{Video recording}
%Four Japanese graduate students (4 males) performed the tasks following the texts while equipped with a head-mounted camera (see \cref{fig:camera}). %While the tasks, they were equipped with a head-mounted camera Panasonic HX-A500 as shown in \cref{fig:camera}. 
%A participant worked on every task two or four times, totaling 10 recordings for each task. To record various mistake actions, they intentionally performed mistake actions that we assumed in advance (\eg, skipping several steps). Out of ten times to record videos, they contained the intentional mistakes five times and tried to follow procedures correctly for the other five times. Note that accidental mistakes due to carelessness also happened even in the latter.
% To avoid the influence of background changes, a desk with objects, tools, and printed procedural texts is placed in the same indoor position across the recording sessions. The participants are recorded sitting to capture manipulated objects in detail.

\begin{figure}[t!b]
    \centering
    \includegraphics[width=0.8\columnwidth]{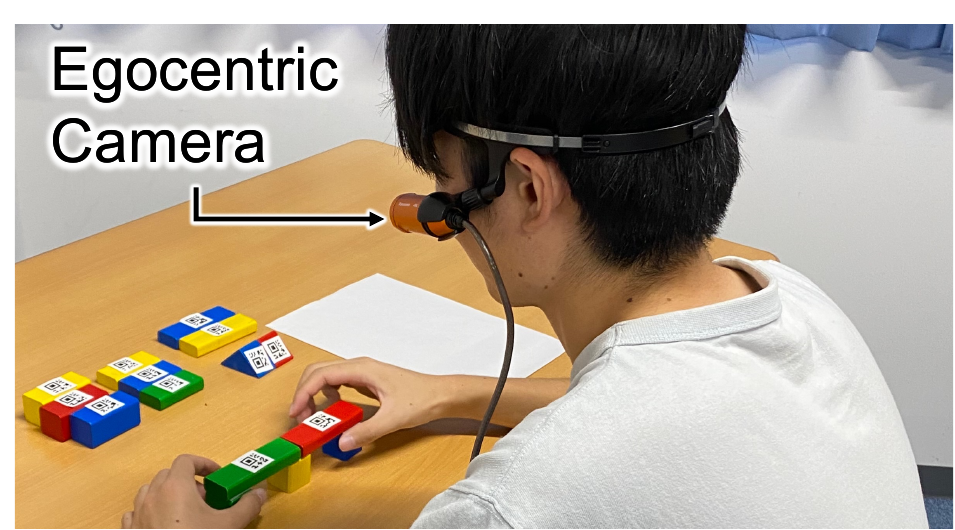}
    \caption{Egocentric head-mounted camera on participants.}
    \label{fig:camera}
\end{figure}

We asked four graduate students to perform the tasks based on the procedural texts. To record their activities, a head-mounted camera (Panasonic HX-A500) was used (\cref{fig:camera}). The egocentric videos were recorded at 30 fps with 4K RGB resolution. We chose the egocentric perspective to capture the participants’ visual attention and fine-grained hand-object interactions, which are critical for modeling procedural understanding. During the recordings of EC and BB, images of the final products were given to the participants.\footnote{Our preliminary experiments showed these tasks were difficult to complete only with the texts.} To avoid the influence of background changes, we fixed the initial locations of the objects, tools, and printed procedural texts. Further, the participants were instructed to work in the chair to capture manipulated objects in detail.

Each participant was asked to perform each task two or four times, totaling 10 recordings per task. For each task, five of 10 recordings contain mistakes that the participants performed intentionally. For the other five recordings, the participants were asked to follow the procedural texts avoiding mistakes. Note that the latter five recordings still contain mistakes due to their careless errors.

%%%%%%%%%%%%%%%%%%%%%
% ICIP2025のサプマテ
%%%%%%%%%%%%%%%%%%%%%

% \noindent\textbf{Participants, camera, and environments.}
% Four Japanese graduate students (4 males) performed the tasks following procedural texts. The participants were equipped with a head-mounted camera Panasonic HX-A500 as shown in \cref{fig:camera}. It is a 30 fps video camera with 4K RGB resolution. 
% When recording the tasks, the participants referred to the Japanese versions of procedural texts. We also gave them images of the finished products in electrical circuits and toy building blocks. This is because our preliminary experiments showed that the two tasks were difficult to complete only with the texts.
% To avoid the influence of background changes, a desk with objects, tools, and printed procedural texts is placed in the same indoor position across the recording sessions. The participants are recorded sitting to capture manipulated objects in detail.

% \noindent\textbf{Recording process.}
% A participant worked on every task 2 or 4 times, totaling 10 recordings for each task.
% To record various mistake actions, they intentionally perform mistake actions that we assume in advance (\eg, skipping several procedures).
% Out of ten times to record videos, they contained the intentional mistakes five times and tried to follow procedures correctly for the other five times. Note that accidental mistakes due to carelessness also happened even in the latter, which is desirable to contain natural mistakes in EgoOops.

%%%%%%%%%%%%%%%%%%%%%

\subsection{Annotations}
\label{subsec:annotation}

% 概要
%EgoOops contains annotations of video-text alignment, mistake labels, and their descriptions. One person annotates the whole dataset using a web annotation tool described in the supplementary materials.
\egooops provides annotations of video-text alignment, mistake labels, and textual descriptions of the mistakes. 

% 手順アノテーション
%We first annotate video-text alignment by extracting start and end timestamps (\ie, segments) and mapping them to the corresponding steps in a procedural text. Note that we also annotate steps not written in the procedural text as \term{undefined} (\eg, grasping the wrong object). The video-text alignment itself reveals order mistakes. For example, if step 2 occurs before step 1, these steps are considered swapped. Thus, we do not explicitly annotate order mistakes.
Video-text alignment refers to the segments in the recorded videos, and each segment is represented as a start and end timestamp. Each segment corresponds to a human action, which is a step in the procedural text or a mistake (\eg, grasping the wrong object).

% 作業誤りアノテーション
%If a segment indicates an execution mistake, we assign one label of the six execution mistake classes listed in \cref{subsec:term}. This categorization can be used to analyze mistake patterns across diverse tasks (\eg, a commonly frequent category of mistakes). In addition, we also provide the mistake segments with descriptions of why the performed actions are considered mistakes. They can be used to analyze model performance in detail or interact with workers through smart assistants.
Mistake labels are assigned to segments that correspond to execution mistakes. A mistake label is one of the six mistake types in \cref{subsec:term}. In addition to the mistake labels, we also provide textual descriptions of why the performed actions are considered mistakes.

% 一致率
%We ensure the quality of the original annotations by testing for agreement with another annotator. For video-text alignment, the annotator newly extracts segments and maps them to the corresponding step labels, resulting in the temporal Intersection over the Union (tIoU) to the original at 88.8. In addition, they watch the original segments and give new mistake labels and descriptions. The labels and descriptions are compared using Cohen's kappa~\cite{cohen_coefficient_1960} and BERTScore~\cite{zhang_bertscore_2020}, achieving 86.8 and 96.3, respectively. These high scores confirm the quality of the original annotations.
We asked two persons for annotation, who are called annotator-A and annotator-B to avoid confusion. Annotator-A was asked to annotate the whole dataset using a web annotation tool that we developed. After the annotation, annotator-B was asked to annotate several videos from the dataset again to calculate inter-annotator agreements. For video-text alignment, annotator-B extracted segments from videos and mapped them to the corresponding step labels. The temporal Intersection over the union (tIoU) to the original segments was then calculated. For mistake labels and descriptions, annotator-B was asked to provide mistake labels and descriptions based on the segments annotator-A annotated. The labels and descriptions between the annotators were compared using Cohen's kappa~\cite{cohen_coefficient_1960} and BERTScore~\cite{zhang_bertscore_2020}. The tIOU was $88.8$, and Cohen's kappa and BERTScore were $86.8$ and $96.3$, respectively, ensuring that the annotations of the dataset are consistent.

\subsection{Statistics}
% 映像とテキストの統計量から，異なる傾向の作業を含むことを示す．
% 誤りアノテーションの統計量から，全作業で共通の誤りや特定の作業に特有の誤りが存在することを示す．また，多様で豊富な誤りを収録したことを示す．
In this section, we first report video- and text-side statistics on EgoOops, then discuss mistake label statistics. The statistics of mistake descriptions are written in the supplementary materials.

% 映像とテキスト
\Cref{tab:video-text-stats} shows different trends between the tasks in terms of video duration, segment duration, and the number of segments. For video duration, the longest is the cardboard task at 26.1 minutes, while the shortest is the building block task at 1.9 minutes.
As for the texts, \cref{tab:video-text-stats} compares the number of steps in a procedural text and the number of words per step. The task with the most steps is the cardboard task, while the task with the longest instructions is the building block task.
These statistics indicate that EgoOops covers a variety of procedural tasks ranging from short to long.

\begin{table}[t!b]
    \centering
    \caption{Statistics of recorded videos and procedural texts.}
    \label{tab:video-text-stats}
    \scalebox{0.85}{
    \begin{tabular}{l|rr|rrr|rr} \toprule
        \multirow{3}{*}{Task} & \multicolumn{2}{c|}{Videos} & \multicolumn{3}{c|}{Segments} & \multicolumn{2}{c}{Texts} \\
         & \#vid. & \makecell{Avg.\\(min)} & \#seg. & \makecell{\#seg.\\/ \#vid.}  & \makecell{Avg.\\(sec)} & \makecell{\#steps\\per text} & \makecell{\#words\\/ \#steps} \\ 
        \midrule
        \electricalcircuit & 10 & 3.2 & 98& 9.8& 15.4 & 8\phantom{.0} & 7.6 \\
        \colormixture & 10 & 4.4 & 91& 9.1& 25.8 & 8\phantom{.0} & 17.0 \\
        \ionicreaction & 10 & 5.4 & 95 & 9.5 & 29.7 & 9\phantom{.0} & 12.7 \\
        \buildingblock & 10 & 1.9 & 87 & 8.7 & 9.0 & 7\phantom{.0} & 18.6 \\
        \cardboard & 10 & 26.1 & 167& 16.7& 86.7 & 14\phantom{.0} & 9.6 \\ \midrule
        All & 50 & 8.2 & 538& 10.8& 40.7 & 9.2 & 13.5 \\ \bottomrule
    \end{tabular}
    }
\end{table}

% 誤りラベル

\Cref{tab:mistake-action-classes} shows the counts of the labels for execution mistakes. In total, \egooops contains 95 execution mistakes. Counting the number of each type of mistake, the two most frequent labels are \term{incorrect-object-picking} (label 2) and \term{wrong-way} (label 5). In addition, we also find unique mistake patterns of each task. For example, \term{accidental-mistakes} (label 4) frequently happen in ionic reaction experiments but rarely in the other tasks. We expect the reason is that the ionic reaction experiments involve moving a small metal piece with tweezers and dropping solution into a narrow space. Overall, many execution mistakes occur in the videos of \egooops, and the tasks have their own frequent mistake types.

\begin{table}[t!b]
\centering
\caption{The number of mistake labels.}
\label{tab:mistake-action-classes}
\scalebox{0.75}{
{\tabcolsep = 1mm
\begin{tabular}{l|rrrrrr} \toprule
\multirow{2}{*}{Task}& \multicolumn{6}{c}{Mistake label}\\
 & 1. \object& 2. \mispick& 3. \correction& 4. \accident& 5. \howto&6. \others\\ \midrule
\electricalcircuit & 9& 5& 1& 2& 3&2\\
\colormixture & 4& 8& 0& 2& 5&3\\
\ionicreaction   & 0& 3& 1& 5& 6&4\\
\buildingblock & 2& 5& 5& 1& 5&1\\
\cardboard & 5& 3& 0& 1& 2&2\\
\midrule
Total& 20& 24& 7& 11& 21&12\\
\bottomrule
\end{tabular}
}
}
\end{table}

%%%%%%%%%%%%%%%%%%%%%
% 修論
%%%%%%%%%%%%%%%%%%%%%

% \paragraph{順序誤り}
% \Cref{tab:mistake-stats-task}に，各作業ごと及び全ての作業について，\cref{subsec:term}で述べた4種類の\ordermistakes 及びその合計の数を示している．全て合わせると，138回の\ordermistakes が生じている．誤りの種類ごとにみると，\term{\resume}が最も多く，合計で51回生じている．次に多く生じているのが，\term{\outoforder}であり，合計で40回である．作業ごとにみると，最も順序誤りが多いのは，段ボール工作の61回である．その中でも上位2種類の誤りは，\term{\resume}と\term{\outoforder}であり，作業者が手順書内の手順を自由な順番で実行していると読み取れる．対照的に，光の混色反応実験とイオン反応実験では，\term{\undefined}が最も多く生じており，余分な手順を挟んでしまっていることが分かる．このように，それぞれの作業は，\ordermistakes に関して，異なる傾向を示している．

% \begin{table}[t!b]
%     \centering
%     \caption{\ordermistakes の数．}
%     \label{tab:mistake-stats-task}
%     \begin{tabular}{l|rrrr|r} \toprule
%         \multirow{2}{*}{作業} & \multicolumn{5}{c}{\ordermistakes の種類}\\
%          & \missing & \outoforder & \resume & \undefined & 合計 \\ \midrule
%         \electricalcircuit & 2 & 10 & 12 &  6 & 30 \\
%         \colormixture      & 1 &  2 &  3 &  9 & 15 \\
%         \ionicreaction     & 2 &  4 &  1 &  6 & 13 \\
%         \buildingblock     & 0 &  3 &  6 & 10 & 19 \\
%         \cardboard         & 7 & 21 & 29 &  4 & 61 \\ 
%         \midrule
%         全て                &12 & 40 & 51 & 35 &138 \\
%         \bottomrule
%     \end{tabular}
% \end{table}

\section{Application: mistake action detection}
% EgoOopsデータセットに対して，Mistake action detection タスクを解く．
% このタスクは，入力映像から作業誤り区間とそのクラスを予測するものである．
% 二つのサブタスクに分解する：まずvideo-text alignment，その出力に対してmistake label classification

We propose a text-oriented approach to mistake action detection. This section formalizes the problem as consisting of video-text alignment and mistake label classification and explains our approach to them.

\subsection{Problem formalization}
\label{subsec:problem-settings}
% Overview
Mistake action detection consists of two problems in our formalization: video-text alignment and mistake label classification. It leads to localizing temporal segments and classes of mistakes in a video.

% Video-text alignment / (procedural) step localization
% 入力：untrimmedな映像と手順書
% 出力：各手順の開始・終了時刻
Given an untrimmed video $\mathbf{V} = (\mathbf{f}_1, \dots, \mathbf{f}_L)$ and the corresponding procedural text $\mathbf{T} = (\mathbf{t}_1, \dots, \mathbf{t}_K)$, our first problem is video-text alignment. Here, $\mathbf{V}$ consists of $L$ frames, and $\mathbf{T}$ includes $K$ steps of instructions. For $k$-th step $\mathbf{t}_k$, our goal is to localize the start and end frame numbers $(s_k, e_k)$.

% Mistake label classification
% 入力：各手順に対応する映像区間
% 出力：その区間の誤りクラス
% 3. correction 以外の誤りを"mistake"という一つのクラスにまとめた
The alignment outputs are passed to the next problem of mistake label classification, which predicts a mistake label for each step segment. The $k$-th step segment is represented by extracted video frames based on the start and end frame numbers as $\mathbf{V}'_k = (\mathbf{f}_{s_k}, \dots, \mathbf{f}_{s_e})$. Our objective is to assign the segment $\mathbf{V}'_k$ one label of mistake classes $(c_1, \dots, c_N)$, where $c_n$ and $N$ represent the name of the $n$-th class and the number of the classes, respectively.  
In our experiments, we group the mistakes except for 3. \term {correction} (see \cref{subsec:term}) into the common \term{mistake} class and solve the classification of three classes: \term{correct}, \term{mistake}, and \term{correction}. \footnote{Our preliminary experiments showed that distinguishing the seven classes (six classes in \cref{subsec:term} plus the \term{correct} class) is difficult currently.}

\subsection{Video-text alignment}
\label{subsec:our-approach}

For video-text alignment, we enhance an existing model of StepFormer~\cite{dvornik_stepformer_2023} by introducing an additional fully supervised loss function, termed StepFormer++. We first provide an overview of the original StepFormer and then explain how we extend it.

\noindent\textbf{Preliminary: StepFormer.} 
% モデルの説明
% StepFormerは映像とテキストのアライメントを，自己教示あり学習するために提案された．
StepFormer was originally proposed for learning video-text alignment from untrimmed videos accompanied by narrations in a self-supervised manner~\cite{dvornik_stepformer_2023}.
% It was trained on HowTo100M~\cite{miech_howto100m_2019} in the original paper, which comprises web video and narration pairs. 
StepFormer is a Transformer decoder~\cite{vaswani_attention_2017} equipped with $U$ learnable queries. Given video features extracted using UniVL~\cite{luo_univl_2020}, the queries attends to the video features, producing $U$ contextualized vectors called \term{step slots}, which capture key steps in the video. The step slots temporally align with narration vectors extracted using UniVL, where they use a sequence-to-sequence alignment algorithm Drop-DTW~\cite{dvornik_drop-dtw_2021}. Considering this alignment as positive pairs, the loss to supervise the step slots is calculated as contrastive one  InfoNCE~\cite{oord_representation_2018} at both local (same video-narration pairs) and global (different video-narration pairs) levels. 
During inference, StepFormer can temporally localize the video segment of each step instruction (\ie, video-text alignment). Specifically, the extracted step slots align with the step instructions in procedural texts, allowing unmatched slots to be dropped. Next, the remaining slots are aligned with the video to identify the start and end times of each step. In these alignment processes, Drop-DTW is used again.

\noindent\textbf{Pre-training.}
% なぜStepFormerなのか
% 自己教示あり学習なため：1. アノテーションの労力を避けながら大規模なデータセットを活用できる, 2. 事前学習によりEgoOopsのデータセットの小ささの悪影響を緩和できる
% 原論文では，ナレーションされたweb上の映像 (HowTo100M) からUniVLで取り出した特徴量で学習していたが，我々は一人称視点の映像 (Ego4D) で自己教示あり学習する
We select StepFormer because its self-supervised learning can be used for pre-training to mitigate the negative impact of the small size of EgoOops. In the original paper~\cite{dvornik_stepformer_2023}, StepFormer is trained on untrimmed web videos and narrations in HowTo100M~\cite{miech_howto100m_2019} using UniVL~\cite{luo_univl_2020} features. Instead, we train it on Ego4D~\cite{grauman_ego4d_2022} using EgoVLPv2~\cite{pramanick_egovlpv2_2023} features to fill the domain gap between web and egocentric videos. Ego4D is a massive-scale egocentric video dataset accompanied by transcriptions~\cite{grauman_ego4d_2022}, hence we can pre-train StepFormer following the same procedure as the original one.

\begin{figure}[t!b]
    \centering
    \includegraphics[width=\columnwidth]{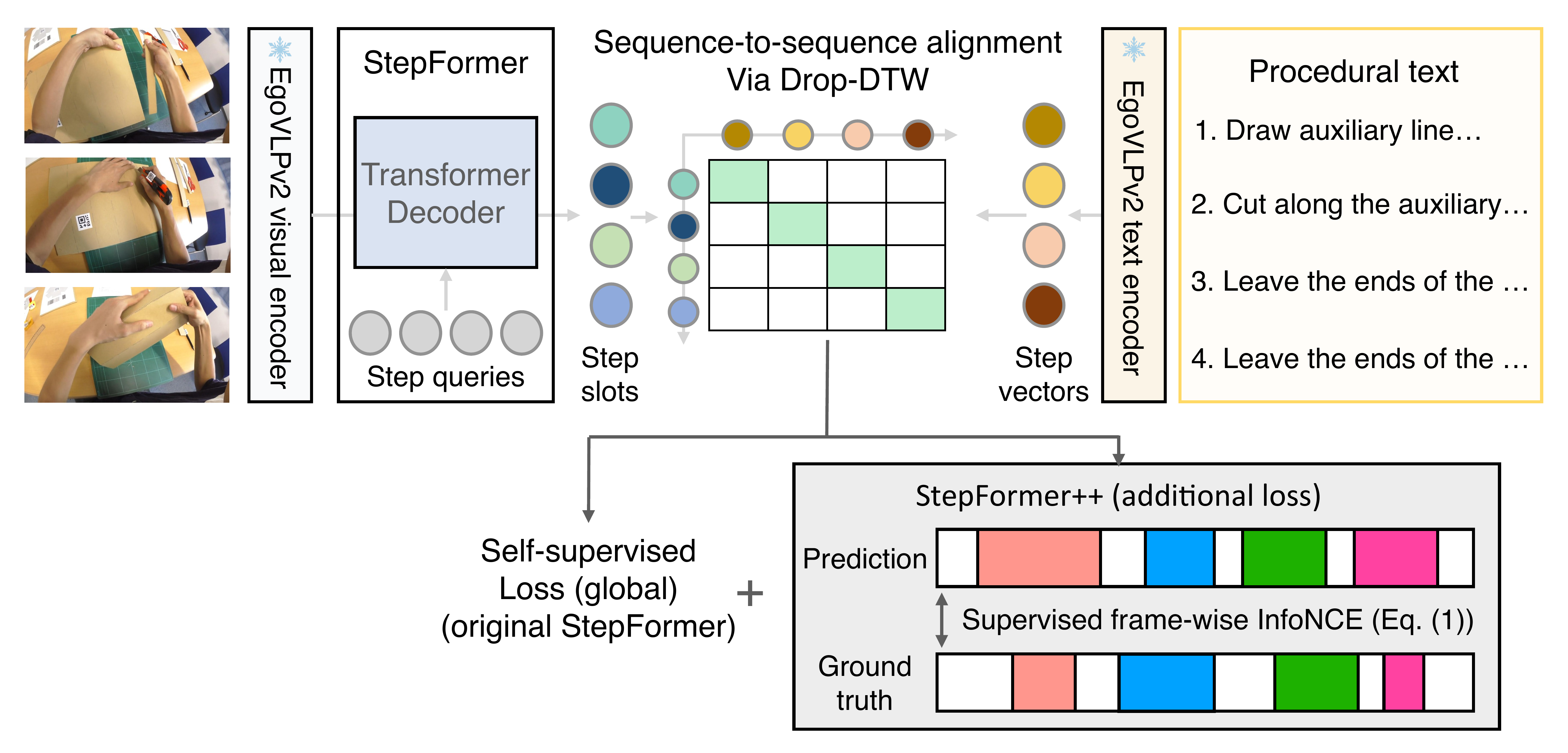}
    \caption{An overview of StepFormer++.}
    \label{fig:stepformer}
\end{figure}

\noindent\textbf{StepFormer++.}
The pre-trained model is fine-tuned on EgoOops with an additional loss to train StepFormer in a fully supervised setting by leveraging the video-text alignment annotations.
\Cref{fig:stepformer} shows an overview of the resulting model StepFormer++. The overall process is the same as the original StepFormer. Given $(\mathbf{V},\mathbf{T})$, the model first extracts video $\mathbf{H}_v=(\mathbf{h}_v^1, \dots, \mathbf{h}_v^l, \dots, \mathbf{h}_v^L)$ and text $\mathbf{H}_t=(\mathbf{h}_t^1, \dots, \mathbf{h}_t^k, \dots, \mathbf{h}_t^K)$ features using EgoVLPv2 instead of UniVL. Then, the Transformer decoder makes slot queries attend $\mathbf{H}_v$, producing step slots $\mathbf{S}=(\mathbf{s}_1, \dots, \mathbf{s}_u, \dots, \mathbf{s}_U)$. Finally, the model acquires the step-text alignment via Drop-DTW by computing a similarity matrix of $\mathbf{S}$ and $\mathbf{H}_t$. The step-text alignment is used to compute the original loss at only the global (different video-narration pairs) level. Instead of the local (same video-narration pairs) loss, we add a new loss to the StepFormer for learning from video-text alignment annotations. After Drop-DTW between the slots and step instructions\footnote{We leave the same number of slots as the steps in the procedural text.}, we supervise the remaining step slots $\hat{\mathbf{S}}=(\hat{\mathbf{s}}_1, \dots, \hat{\mathbf{s}}_k, \dots, \hat{\mathbf{s}}_K)$ to match with the start and end frame numbers $(s_k,e_k)$. Specifically, this is calculated using the InfoNCE framework:
\begin{equation}
    \mathcal{L}_{\mathrm{supervised}} (\hat{\mathbf{s}}_k, \mathbf{H}_v) = - \log \frac{\sum_{j \in [s_k, e_k]} f(\hat{\mathbf{s}}_k, \mathbf{h}^j_v)}{\sum_l f(\hat{\mathbf{s}}_k, \mathbf{h}^l_v)},
\end{equation}
where $f(\hat{\mathbf{s}}_k, \mathbf{h}^*_v) \! = \! \exp (\cos (\hat{\mathbf{s}}_k, \mathbf{h}^*_v)) / \gamma$, and $\gamma$ is a scaling temperature. We add this loss to the original ones, and the pre-trained model is fine-tuned on EgoOops.

\subsection{Mistake label classification}
\label{subsec:our-classifier}

\noindent\textbf{Multi-modal classifier.}
Given a pair of predicted video segment $\mathbf{V}'$ and $k$-th step instruction $\mathbf{t}_k$, the model predicts the mistake label. Specifically, the model first convert $\mathbf{V}'$ and $\mathbf{t}_k$ into video $\mathbf{H}_v'=(\mathbf{h}_v^{s_t}, \dots, \mathbf{h}_v^{e_t})$ and text $\mathbf{h}_t^k$ features using EgoVLPv2. Then, it computes the mean of $\mathbf{H}_v'$, concatenates the averaged vector with $\mathbf{h}_t^k$, and forwards it into a two-layer perceptron $g$ with ReLU function as following: $\mathbf{z}_k = g(\mathrm{concat}(\mathrm{mean}(\mathbf{H}_v'), \mathbf{h}_t^k))$, where $\mathbf{z}_k = (z_{k}^1, \dots, z_k^n, \dots, z_k^N)$ represents the logits for classes, and $N$ is the number of the classes. The model applies the argmax operation on $\mathbf{z}_k$ and outputs the prediction label.

\noindent\textbf{Training.}
To train the model, we use the class-balanced focal loss~\cite{cui_class-balanced_2019} because the frequency of the \term{mistake} and \term{correction} labels is lower than the \term{correct} label.
Specifically, using $\mathbf{z}_t$, the loss is calculated as following:
\begin{equation}
    \mathcal{L}_{\mathrm{classification}} (\mathbf{z}_t) = - \frac{1 - \beta}{1 - \beta^{r_{c_n}}} \log \frac{\exp (z_t^{c_n})}{\sum_j \exp(z_{t}^{c_j})},
\end{equation}
where $r_{c_n}$ is the number of training samples belonging to the class $c_n$, and $\beta \in [0,1)$ is a hyperparameter.
We adopt teacher forcing~\cite{graves_generating_2014,vaswani_attention_2017} as the training strategy. Specifically, we input the ground-truth segment of the $t$-th step to the model to stabilize training whereas the predicted ones are used for testing.

\subsection{Implementation details.}
We follow the official implementation of StepFormer and use the same hyperparameters as stated in \cite{dvornik_stepformer_2023} unless we mention modifications.
We set the number of step queries to be $U=32$ and the batch size to be 6 for fine-tuning StepFormer++ on EgoOops. The video and text feature dimension of EgoVLPv2 is $d=4,096$.
We use Drop-DTW with an 80 percentile drop cost~\cite{dvornik_drop-dtw_2021} to align the step slots and video features.
We set $\gamma = 0.03$ in the InfoNCE loss and $\beta = 0.9999$ in the class-balanced loss.
For the mistake label classification, we train the classifier in 1,200 epochs.

\section{Experiments}

We first report an end-to-end performance on mistake action detection in Section \ref{subsec:experiments_mistake_action_detection}. We then conduct in-depth experiments on video-text alignment and mistake label classification individually in Section \ref{subsec:exp-alignment} and \ref{subsec:exp-classification}.

\subsection{Mistake action detection}
\label{subsec:experiments_mistake_action_detection}

\noindent\textbf{Baseline.}
We do not adopt existing mistake detection methods as baselines because they do not fit our settings. For instance, Assembly101~\cite{sener_assembly101_2022} and CaptainCook4D~\cite{peddi_captaincook4d_2024} assume trimmed video clips as inputs, while our task assumes untrimmed videos. The method in \cite{seminara_differentiable_2024-1} focuses on ordering mistakes in the videos and does not address execution mistakes.
EgoPED~\cite{lee_error_2024} and AMNAR~\cite{huang_modeling_2025} are the closest to our setting as they predict both segments and mistake labels. However, their methods are based on anomaly detection, predicting binary labels of ``correct'' and ``mistakes,''; thus they cannot predict the three classes of ``correct,'' ``mistakes,'' and ``correction.''

Therefore, instead of existing mistake action detection models, we compare our method with the recent temporal action localization (TAL) model, ActionFormer~\cite{zhang_actionformer_2022}.
This is because TAL operates under similar conditions, where the models detect both temporal segments and their action labels. In our experiments, we train ActionFormer to predict mistake labels for the detected segments, instead of action labels as in the original settings. For a fair comparison, we apply NMS to retain as many segments as the ground truths. In addition, since our metrics require step labels, we assign them to the segments in order from the start to the end of the videos. Note that the inputs for TAL are videos only.

\noindent\textbf{Evaluation metrics.}
We follow TAL~\cite{zhang_actionformer_2022} and report mean average precision (mAP) at tIoU thresholds of 0.1, 0.2, and 0.3.
It computes the mean of average precision across only \term{mistake} and \term{correction} classes because we focus on mistake detection performance. 
Since our problem formalization involves video-text alignment (see \cref{subsec:problem-settings}, our metrics require correctly predicting both step and mistake labels.
Note that ActionFormer processes only videos and does not conduct video-text alignment~\cite{zhang_actionformer_2022}; yet we assign step labels to the segments sequentially from the video's start to end for fair comparison.

\noindent\textbf{Splits.}
Our EgoOops dataset is relatively small compared to other action mistake datasets. To ensure reliable results, we perform 5-fold cross-validation. We divide the 50 videos into a 30/10/10 split for training, validation, and testing, respectively. All 30 training videos from the five tasks are used to train one unique model for scalability and generalization across diverse tasks. We report the average test-set scores using the model weights that achieve the highest performance on the validation set. To construct folds, we pay attention to the two points. First, each validation and testing fold contains one correct and one mistake video for every task, totaling 10 videos. Second, each fold consist of the same workers' videos as following the group k-fold~\cite{scikit_learn_31_nodate}.
This allows us to test the models on unseen worker's activities, minimizing the bypass possibility to learn the worker-specific features to detect mistakes.

\begin{figure}[t!b]
    \centering
    \includegraphics[width=0.9\linewidth]{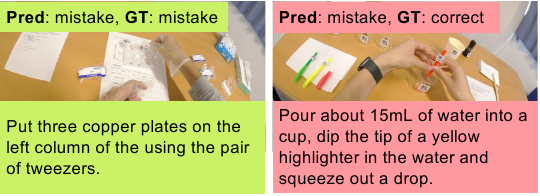}
    \caption{Success (left) and failure (right) cases of mistake detection. Text boxes at the bottom show the predicted steps.}
    \label{fig:res-qualitative-mistake}
\end{figure}

\noindent\textbf{Results.}
% 誤り検出の定量結果
\Cref{tab:res-mistake-action-detection} shows the performance on mistake action detection. Our proposed method achieves an average mAP of 2.5, surpassing ActionFormer's 0.7. In contrast, our score is still far from the oracle's 34.7, which uses ground-truth step segments and only performs mistake label classification. This suggests that accurate video-text alignment largely improves mistake action detection.
% 定性結果
In addition, we explore the success and failure examples as shown in \cref{fig:res-qualitative-mistake}. In the success case, the correct alignment leads to the finding of a mistake; in the failure case, the localized step is incorrect, hurting the performance of mistake classification. 
% 分類器からのテキストのablation
Moreover, we conduct an ablation study on input modalities, as shown in \cref{tab:res-ablation-classifier-modalities}. When comparing models with and without text input, we observe that the average mAP drops significantly from 2.5 to 0.3. This highlights the importance of textual information for accurate mistake classification.

\begin{table}[t!b]
    \centering
    \caption{Results of mistake action detection. Oracle denotes upper-bound performance using ground-truth step segments.
    }
    \label{tab:res-mistake-action-detection}
    \begin{tabular}{l|rrrr} \toprule
        \multirow{2}{*}{Methods} & \multicolumn{4}{c}{mAP@tIoU} \\
         & 0.1 & 0.2 & 0.3 & Avg. \\ \midrule
        ActionFormer~\cite{zhang_actionformer_2022} & 1.8 & 0.1 & 0.1 & 0.7 \\
        % ActionFormer (two-stage) ~\cite{zhang_actionformer_2022} & & & & \\
        % TriDet~\cite{shi_tridet_2023} & & & & \\
        % TriDet (two-stage) ~\cite{shi_tridet_2023} & & & & \\
        StepFormer++ w/ MLP (ours)& 2.6& 2.5& 2.5& \textbf{2.5}\\ \midrule
        GT steps (oracle) &  \multicolumn{4}{c}{34.7}\\ \bottomrule
    \end{tabular}
\end{table}

\begin{table}[t!b]
    \centering
    \caption{Ablation study of textual inputs to the mistake label classifier of our approach.}
    \label{tab:res-ablation-classifier-modalities}
    \scalebox{0.95}{
    \begin{tabular}{cc|rr} \toprule
        \multicolumn{2}{c|}{Inputs to classifier} & \multirow{2}{*}{Avg. mAP} & \multirow{2}{*}{Avg. mAP (oracle)} \\
        Video & Text & & \\ \midrule
        $\checkmark$ &              & 0.3 & 4.7\\
        $\checkmark$ & $\checkmark$ & \textbf{2.5}& \textbf{34.7}\\ \bottomrule
    \end{tabular}
    }
\end{table}

\subsection{Video-text alignment}
\label{subsec:exp-alignment}

% StepFormer++とオリジナルの比較
We conduct detailed experiments on the video-text alignment component. In this experiment, we evaluate the performance of our proposed StepFormer++ against two versions of the original StepFormer~\cite{dvornik_stepformer_2023}. One is trained on Ego4D and evaluated in a zero-shot manner (ZS), while the other is further fine-tuned on EgoOops using a self-supervised approach (SS).

As shown in \cref{tab:res-video-text-alignment}, we report frame-wise F1, Precision, Recall, and Mean over Frames (MoF), following prior work~\cite{shen_learning_2021,dvornik_stepformer_2023}.
Our StepFormer++ achieves an F1 score of 28.1, surpassing both the ZS (24.1) and SS (26.1) variants. These results demonstrate that our fully supervised loss effectively trains StepFormer++ when alignment annotations are available.

\begin{table}[t!b]
    \centering
    \caption{Results of video-text alignment. ZS: zero-shot, SS: self-supervised, FS: fully supervised.}
    \label{tab:res-video-text-alignment}
    \begin{tabular}{l|rrrr} \toprule
        Methods & F1 & Prec. & Rec. & MoF \\ \midrule
        % ActionFormer~\cite{zhang_actionformer_2022} & 11.3 & \textbf{26.0} & 7.5 & 14.6 \\
        % % ActionFormer (two-stage) ~\cite{zhang_actionformer_2022} & & & & \\
        % TriDet~\cite{shi_tridet_2023} & & & & \\
        % % TriDet (two-stage) ~\cite{shi_tridet_2023} & & & & \\
        % Ours & \textbf{23.9}& 24.3& \textbf{23.6} & \textbf{24.5}\\ \bottomrule
        StepFormer (ZS)~\cite{dvornik_stepformer_2023}& 24.1& 24.6& 23.7& 24.9\\
        StepFormer (SS)~\cite{dvornik_stepformer_2023}& 26.1& 26.6& 25.7& 27.0\\
        StepFormer++ (ours, FS)& \textbf{28.1}& \textbf{28.4}& \textbf{27.9}& \textbf{28.1}\\
        \bottomrule
    \end{tabular}
\end{table}

\begin{figure}[tb]
   \centering
   \includegraphics[width=0.8\linewidth]{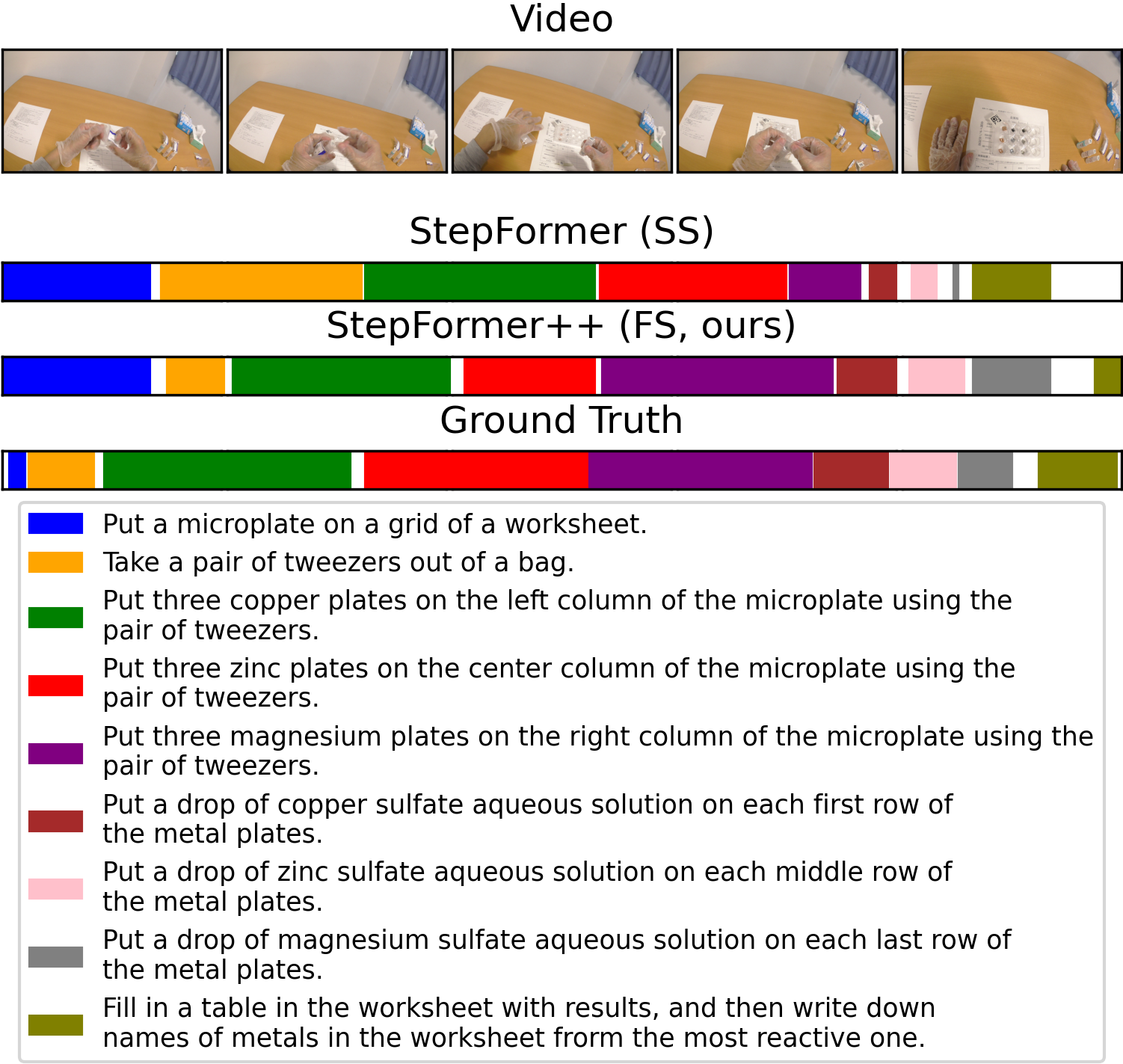}
   \caption{Qualitative results of video-text alignment.}
   \label{fig:res-qualitative-alignment}
\end{figure}

\Cref{fig:res-qualitative-alignment} shows example results of the self-supervised StepFormer and StepFormer++. While the self-supervised one fails to localize the steps of putting copper, zinc, and magnesium plates, our StepFormer++ correctly finds their alignments. This implies that it is difficult to distinguish steps involving similar-looking objects in a self-supervised learning, but our fully-supervised loss helps to learn them. Also, this demonstrates that our fully supervised loss improves StepFormer's video-text alignment capability, thus StepFormer++ achieves more precise step prediction.

%%%%%%%%%%%%%%%%%%%%%

\subsection{Mistake label classification}
\label{subsec:exp-classification}

We address the task of mistake label classification, which predicts one of three labels (correct, mistake, and correction) based on ground-truth segments. As baselines, we evaluate mistake classifiers trained on existing mistake action datasets and report their performance on the EgoOops dataset. Furthermore, we explore the capabilities of recent multi-modal large language models (MLLMs) to assess how accurately they can predict mistake labels.

\noindent\textbf{Classifiers trained on existing datasets.}
% 目的：~~既存データセットとは異なるドメインをEgoOopsで収録していることを示す~~ → 既存データセットで学習したモデルの、未知のドメインへの汎化性能を調べる
% Unlike previous datasets~\cite{sener_assembly101_2022, peddi_captaincook4d_2024}, EgoOops dataset contains a diverse set of unique domains.
% To examine the domain gap, we evaluate models trained on existing datasets using EgoOops dataset in a zero-shot manner.
The EgoOops dataset introduces tasks from a diverse range of previously unexplored domains (see \cref{subsec:task-selection}), beyond those covered in existing assembly~\cite{sener_assembly101_2022,ghoddoosian_weakly-supervised_2023,wang_holoassist_2023,schoonbeek_industreal_2024} and cooking~\cite{lee_error_2024,peddi_captaincook4d_2024} datasets.
To evaluate whether models trained on existing datasets can generalize to unseen domains, we apply these models to predict mistake labels on the EgoOops dataset in a zero-shot manner.

Specifically, we compare the performance of our classifier trained on EgoOops with models trained on Assembly101~\cite{sener_assembly101_2022} and CaptainCook4D~\cite{peddi_captaincook4d_2024}, which are representative benchmarks for assembly and cooking errors, respectively. For Assembly101, we adopt the TempAgg model~\cite{sener_temporal_2020}, a long-range video recognition architecture leveraging TSM features~\cite{lin_tsm_2019}. For CaptainCook4D, we utilize their best-performing model: a multi-layer perceptron (MLP) head on top of a frozen 3D-ResNet backbone~\cite{hara_learning_2017}. In contrast, we train a MLP-based model (see \cref{subsec:our-classifier}) on EgoOops to evaluate the effect of tuning to its domains.

\noindent\textbf{MLLMs.}
% 目的：Mistake descriptionの利用方法を示す．既存のMLLM/VLMでは，高い精度で解けないことを示す．
Finding mistake actions is a visual reasoning task, where models must understand both the video and the associated procedural text to determine whether a worker correctly follows the instructions. MLLMs perform well in visual reasoning benchmarks~\cite{chen_expanding_2024,wang_qwen2-vl_2024}, thus we instruct them to solve mistake label classification given a trimmed video clip, the task’s procedure, and the performed step.

Specifically, we evaluate two leading open-source MLLMs on EgoOops in a zero-shot manner: InternVL2.5-8B~\cite{chen_expanding_2024} and Qwen2-VL-7B-Instruct~\cite{wang_qwen2-vl_2024}. For each instance, we construct the input prompt using a fixed template (\cref{fig:mllm-prompt-template}) designed to 
\begin{enumerate*}[label=(\arabic*)]
    \item provide the full procedure as context,
    \item encourage active identification of mistakes and corrections,
    \item and follow a multiple-choice question format, which MLLMs are well-trained to handle.
\end{enumerate*}
The completed prompt and the video frames are passed to the model, which outputs its answer in free-form text.
Video frame sampling follows each model's official pre-processing: InternVL2.5 takes 24 frames as input, while Qwen2-VL takes 48 frames.

\begin{figure}[t!b]
\centering
\small
\begin{tcolorbox}
Procedure:\\
\verb|{PROCEDURE}|\\
This step: \verb|{STEP_INSTRUCTION}|\\
\\
It is an egocentric video clip where the worker performs an activity referring to the procedure. Note that if the step is "UNDEFINED", it is an extra step not written in the procedure.\\
Carefully look at the clip. Try to find worker's failures of precisely carrying out the step instruction (i.e. mistake) or correction of mistakes. We penalize more for overlooking mistake and correction classes. Select the best option to the following multiple-choice question based on the video clip.\\
\\
Question: Which label best matches the activity performed by the camera wearer?\\
0. correct\\
1. correction\\
2. mistake\\
The best option:
\end{tcolorbox}
\caption{The prompt template for MLLMs. We replace the placeholders with the task's procedure and the performed step for each trimmed video clip.}
\label{fig:mllm-prompt-template}
\end{figure}

%%%%%%%%%%%%%%%%%%%%%
% 修論の記述
%%%%%%%%%%%%%%%%%%%%%

% \subsubsection{各クラスの誤りに対する提案分類器の性能評価}

% \Cref{tab:res-mistake-label-classification-fine-grained}に，6クラスの誤りそれぞれについて，提案分類器によるrecallを示している．スコアが上位の3クラスは，\mispick（クラス2），\correction（クラス3），\accident（クラス4）で，それぞれのrecallは，83.3，57.1，54.5である．一方で，\object（クラス1）と\howto（クラス5）については，それぞれrecallが，35.0と28.6にとどまっている．この二つのクラスの誤りでは，用いた物体や行った動作が手順書通りではないため，物体検出器や行動認識器を提案手法と組み合わせることで，更なる性能向上が期待できる．

% \begin{table}[t!b]
%     \centering
%     \caption{6クラスの誤りそれぞれに対する提案分類器のrecall．3クラス分類で学習したモデルを，6クラスの誤りに対して評価していることに，注意されたい．}
%     \label{tab:res-mistake-label-classification-fine-grained}
%     \begin{tabular}{l|rrrrrr} \toprule
%         手法 & 1. \object & 2. \mispick & 3. \correction & 4. \accident & 5. \howto & 6. \others \\
%         \midrule
%         MLP（提案） & 35.0 & 83.3 & 57.1 & 54.5 & 28.6 & 33.3 \\ 
%         \bottomrule
%     \end{tabular}
% \end{table}

\noindent\textbf{Results.}
\Cref{tab:res-mistake-label-classification} presents the results of mistake label classification. Among the classifiers, the model trained on CaptainCook4D performs better than the uniform sampling baseline, demonstrating a certain level of domain-generalization ability. In contrast, the model trained on Assembly101 does not surpass the baseline, indicating its limited transferability to unseen domains.
The model trained on EgoOops significantly outperforms uniform sampling, highlighting the benefits of domain-specific adaptation.

In terms of MLLMs, Qwen2-VL-7B-Instruct exceeds the fully-supervised MLP classifier when comparing their performance to recognize the ``mistake'' class, suggesting its strong capabilities to find mistakes. However, its performance on recognizing corrections of mistakes is considerably lower compared to the fully-supervised model. 
This gap suggests that current MLLMs have limited ability to reason about whether an action constitutes a correction.

%%%%%%%%%%%%%%%%%%%%%
% 修論の記述
%%%%%%%%%%%%%%%%%%%%%

% \Cref{tab:res-mistake-label-classification-mllm}に，MLLMsによる誤りラベル分類の結果を示している．Qwen2-VL-7B-Instructは，\term{誤り}のクラスについては，43.5と高いF1スコア記録しているが，\term{訂正}のクラスについては，僅か4.4のF1スコアに過ぎない．誤りの訂正を認識するためには，より前の時点で起きた誤りとしてあり得るものを想定し，そこからの復帰動作に現時点の行動が該当するかを予測するという，より複雑な推論が求められる．そのため，この結果は，現時点のMLLMsの推論能力の限界を示唆している．

\begin{table}[t!b]
    \centering
    \caption{Results of mistake action classification. ZS: zero-shot, FS: fully-supervised. Note that \cite{peddi_captaincook4d_2024} addresses binary classification of \term{correct} or \term{mistake}, thus the scores for \term{correction} are not available.}
    \label{tab:res-mistake-label-classification}
    \scalebox{0.71}{
    \begin{tabular}{ll|rrr|rrr} \toprule
        \multicolumn{2}{c|}{\multirow{2}{*}{Methods}} & \multicolumn{3}{c|}{Mistake} & \multicolumn{3}{c}{Correction} \\
         & & Prec. & Rec. & F1 & Prec. & Rec. & F1 \\ 
        \midrule
        \multirow{4}{*}{\rotatebox{90}{Classifier}} & Uniform sampling & 16.4 & 33.3 & 21.9 & 1.3 & 33.3 & 2.5\\
        % \cline{2-8}
         & Assembly101~\cite{sener_assembly101_2022} (ZS) & 14.8 & 14.8 & 14.8 & 2.4 & 42.9 & 4.5 \\
         & CaptainCook4D~\cite{peddi_captaincook4d_2024} (ZS) & 16.5 & 76.1 & 27.1 & \multicolumn{3}{c}{N/A} \\
        % \cline{2-8}
         & MLP (ours, FS) & 35.0 & 48.9 & \textbf{40.8} & 57.1 & 57.1 & \textbf{57.1} \\
        \midrule
        % \multirow{3}{*}{\rotatebox{90}{MLLM}} & GPT-4o~\cite{openai_gpt-4o_2024} & & & & & &\\
        % \cline{2-8}
        \multirow{2}{*}{\rotatebox{90}{{\footnotesize MLLM}}} & InternVL2.5-8B~\cite{chen_expanding_2024} & 47.6 & 34.1 & 39.7 & 0.0 & 0.0 & 0.0 \\
         & Qwen2-VL-7B-Instruct~\cite{wang_qwen2-vl_2024} & 75.0 & 30.7 & \textbf{43.5} & 2.6 & 14.3 & \textbf{4.4} \\
        \bottomrule
    \end{tabular}
    }
\end{table}

\section{Conclusion}

% Short version.
This paper introduced EgoOops dataset that consists of egocentric videos, procedural texts, and three types of annotations: video-text alignment, mistake labels, and mistake descriptions. Based on this, we proposed a text-oriented approach to the task of mistake action detection. Our experiments demonstrated that textual information plays a crucial role in accurately identifying mistakes. Furthermore, we conducted an in-depth analysis of video-text alignment and mistake label classification.
The results revealed that while MLLMs exhibit promising performance in mistake recognition but still struggle to reason about mistake corrections in videos, highlighting a key area for future improvement.

\noindent\textbf{Acknowledgments.}
This work was supported in part by JSPS KAKENHI Grant Numbers 25K21274.

% % Long version.
% This paper proposed the EgoOops dataset that consists of egocentric videos of workers, procedural texts describing tasks across diverse domains, and their annotations. The annotations consist of video-text alignment, mistake labels, and descriptions. The mistake labels revealed both common and unique mistake patterns across the tasks. 
% We also proposed an approach leveraging procedural texts to the problem of mistake detection. Our approach features a combination of video-text alignment and mistake label classification. In experiments on EgoOops, we revealed that our approach outperforms a video-only baseline, and ablating text input decreases our performance, demonstrating the necessity of texts in mistake detection. 
% In addition, we tested existing mistake classifiers and multi-modal large language models (MLLMs) on mistake label classification. The evaluation results showed that mistakes in EgoOops are challenging for the existing classifiers to recognize, suggesting that our new domains lead to novel mistakes. We found that MLLMs cannot accurately find actions to correct mistakes, indicating limited visual reasoning capability.
% % Future work with EgoOops includes improving the performance on both video-text alignment and mistake action detection. We currently do not use QR Codes, but they could be instrumental in achieving this by incorporating object names into the model.
% We release EgoOops and hope to accelerate future studies leveraging text information in mistake detection.

\clearpage
{
    \small
    \bibliographystyle{ieeenat_fullname}
    \bibliography{250701_ICCV_SAUAFG}
}

\clearpage
\setcounter{page}{1}
\maketitlesupplementary

\section{Details of dataset construction}
\label{sec:supp-dataset-construction}

\subsection{Mistake definition}
\label{subsec:supp-term}
% 作業誤りの定義
We define \term{mistakes} as ``deviations from instructional steps.'' This definition leads to two types of mistakes: \term{\ordermistakes} and \term{\executionmistakes}. 

Order mistakes occur when there are discrepancies between the steps executed by a worker and the ones outlined in the procedural text. These include skipping necessary steps (\term{\missing}), swapping the step order (\term{\outoforder}), pausing a step and resuming after others (\term{\resume}), and inserting extra steps (\term{\undefined}).

Execution mistakes occur when a worker fails to follow the instructions while carrying out a step. We categorize them into the following six classes: 
\begin{enumerate}[nosep]
    \item \textbf{Incorrect-object-use (\term{\object})} executes a step with a different object specified in the text. This includes cases when the incorrect number of objects is used.
    \item \textbf{Incorrect-object-picking (\term{\mispick})} picks up an incorrect object, but the worker recognizes the mistake and releases the object. This mistake does not execute a step, unlike incorrect-object-use.
    % \item \textbf{Self-correction} (\term{correction}) executes a step in a wrong way and recognizes and corrects the mistake.
    \item \textbf{Self-correction (\term{\correction})} recognizes and corrects mistakes in a step that has been executed in a wrong way.
    \item \textbf{Accidental-mistakes (\term{\accident})} causes accidental happenings mainly due to carelessness.
    \item \textbf{Wrong-way (\term{\howto})} picks a correct object but executes a step in a way that misaligns with the instruction in the text.
    \item \textbf{Other-mistakes (\term{\others})} induces other types of mistakes. This also includes cases where multiple types of the above mistakes occur simultaneously.
\end{enumerate}

In this study, we aim to detect \executionmistakes by referring to the procedure manual. However, the EgoOops dataset also includes \ordermistakes. These annotations are also added in the hope that they will be used in future studies.

\subsection{Task selection}

\begin{figure*}[t!b]
    \centering
    \begin{subfigure}{0.33\linewidth}
        \centering
        \includegraphics[width=\columnwidth]{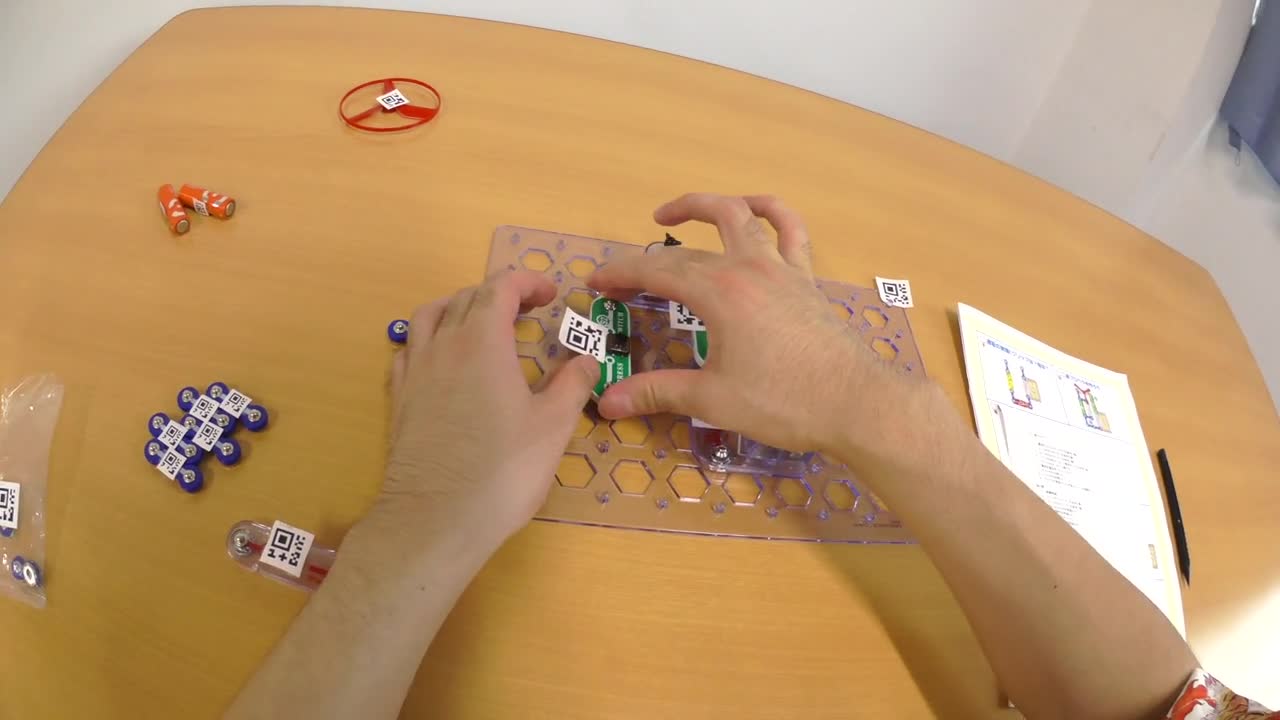}
        \caption{Electrical circuits.}
    \end{subfigure}
    \hfill
    \begin{subfigure}{0.33\linewidth}
        \centering
        \includegraphics[width=\columnwidth]{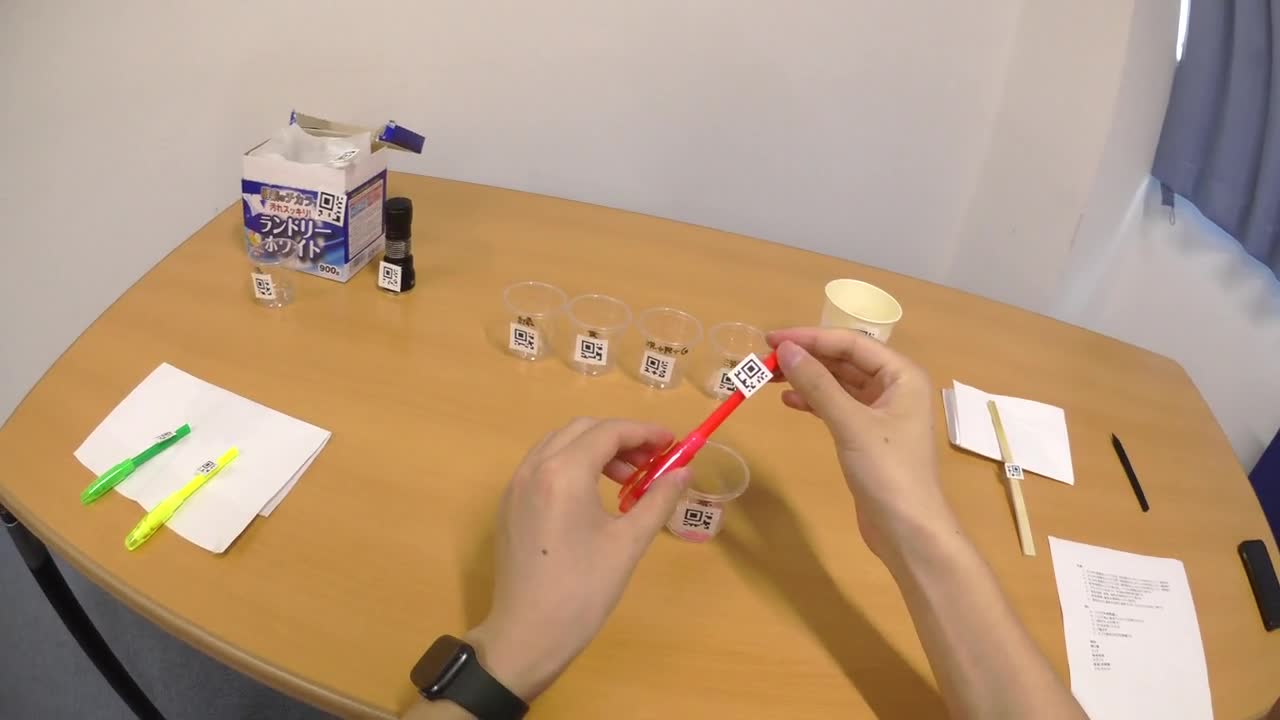}
        \caption{Color mixture experiments.}
    \end{subfigure}
    \hfill
    \begin{subfigure}{0.33\linewidth}
        \centering
        \includegraphics[width=\columnwidth]{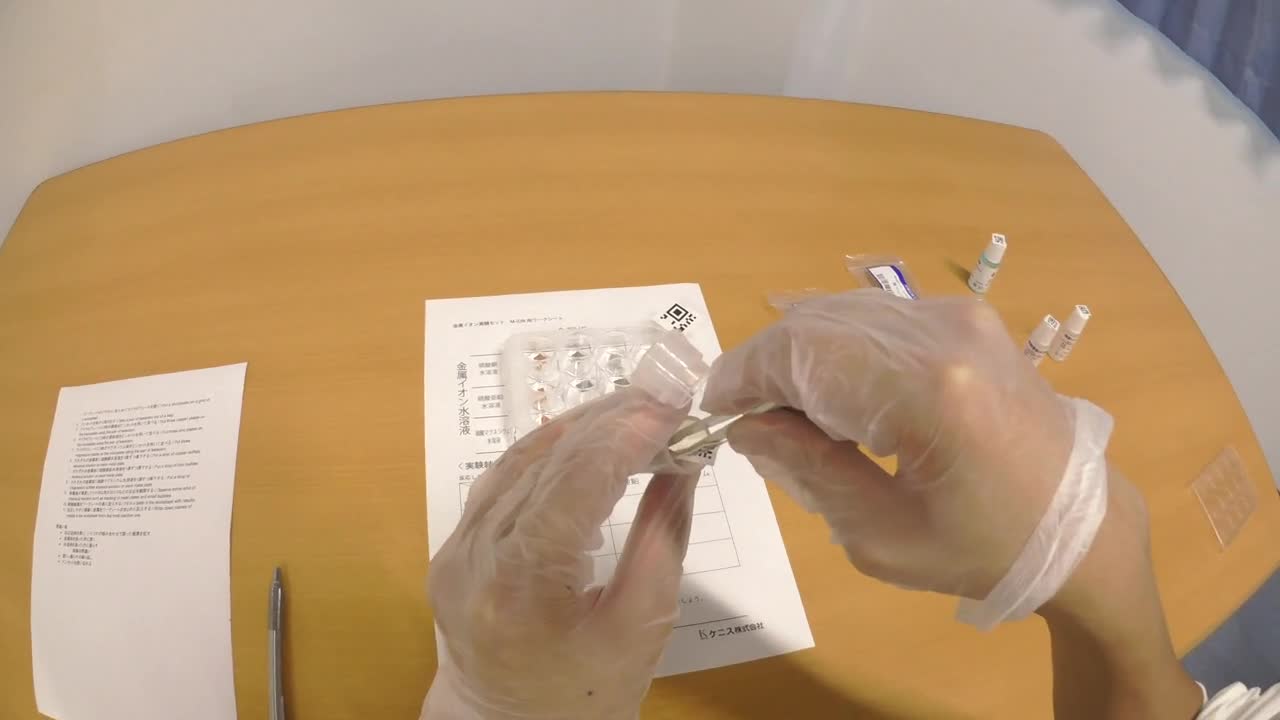}
        \caption{Ionic reaction experiments.}
    \end{subfigure}
    \\ \vspace{2mm}
    \begin{subfigure}{0.33\linewidth}
        \centering
        \includegraphics[width=\columnwidth]{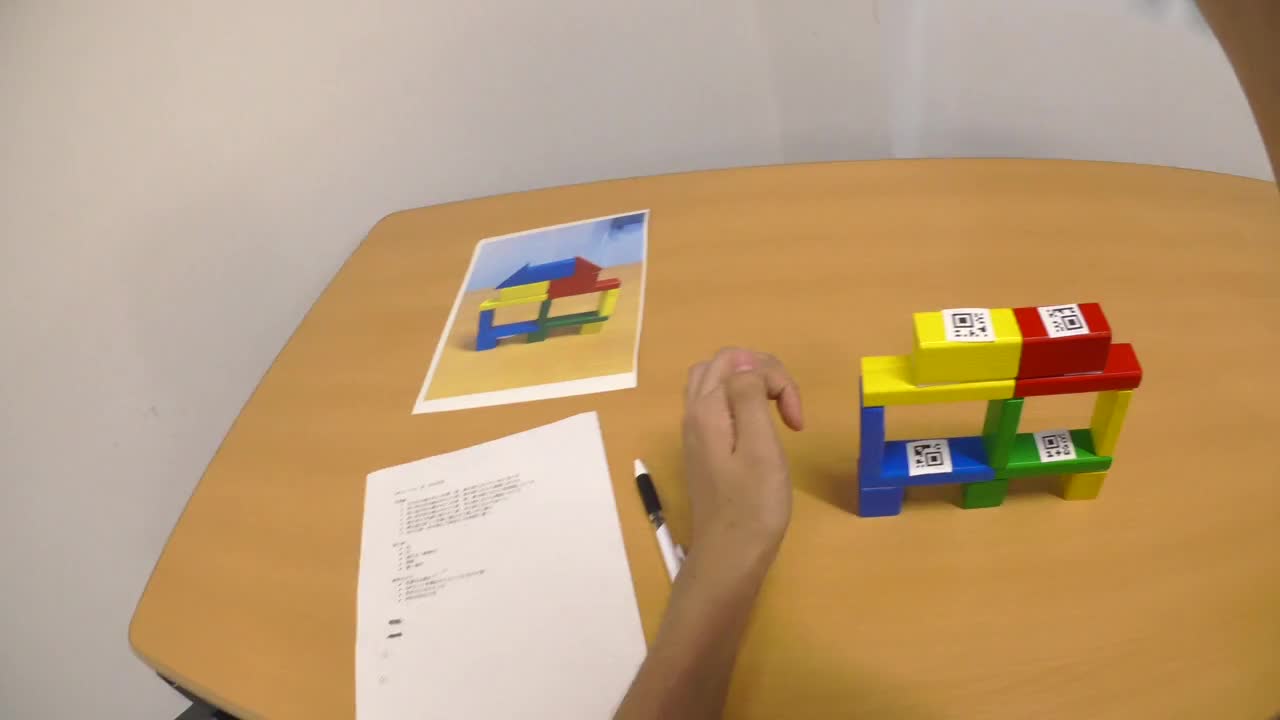}
        \caption{Toy building blocks.}
    \end{subfigure}
    \begin{subfigure}{0.33\linewidth}
        \centering
        \includegraphics[width=\columnwidth]{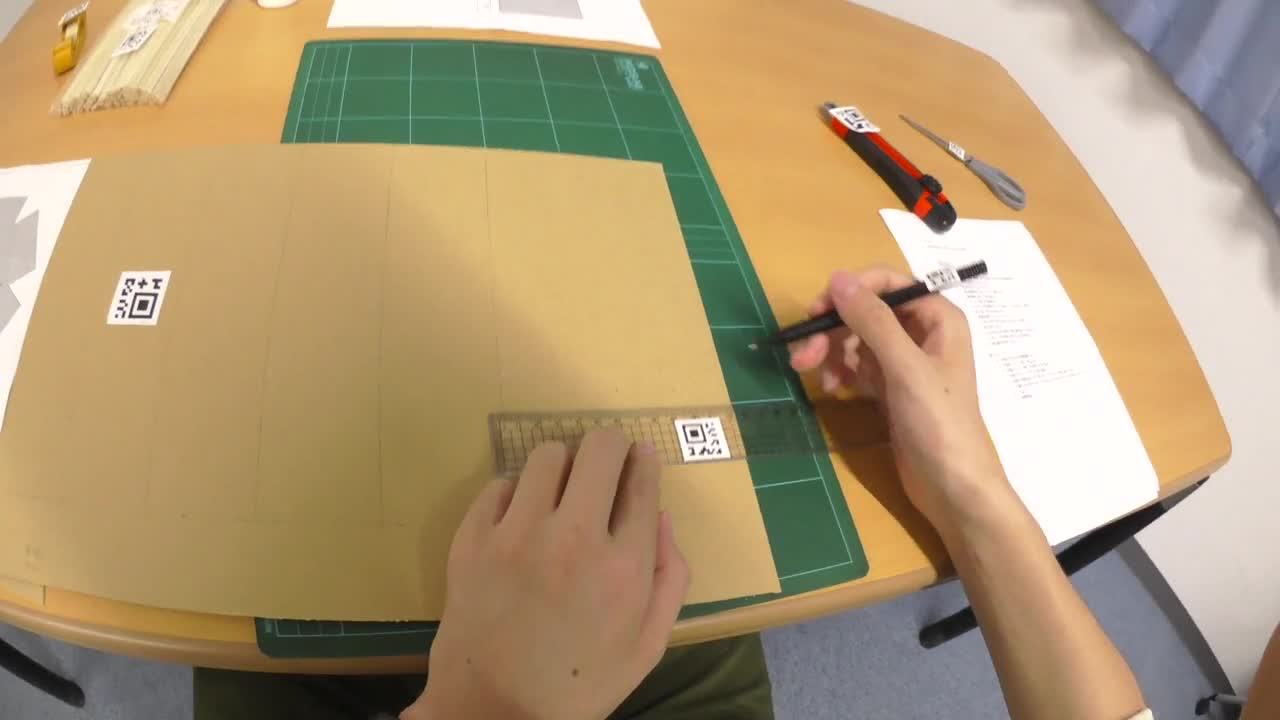}
        \caption{Cardboard crafts.}
    \end{subfigure}
    \caption{Example frames of the five tasks.}
    \label{fig:task-frames}
\end{figure*}

We aim to record procedural activities with mistake actions while following procedural texts in diverse domains.
In addition, we expect to collect various categories of mistakes (\eg, unintentional action, working in the wrong way).
Based on these criteria, we selected the following five tasks (example frames in \cref{fig:task-frames}). 
\begin{itemize}[nosep]
     \item \textbf{Electrical circuits} (\textbf{\term{\electricalcircuit}}): Connecting electrical elements to complete an electrical circuit that turns a propeller.
     \item \textbf{Color mixture experiments} (\textbf{\term{\colormixture}}): Examining the color of various solutions of detergent and fluorescent paint when illuminating them with a blacklight. 
     \item \textbf{Ionic reaction experiments} (\textbf{\term{\ionicreaction}}): Examining ionic reaction by dropping chemical solutions to metal plates.
     \item \textbf{Toy building blocks} (\textbf{\term{\buildingblock}}): Piling up building blocks to construct the specified structure.
     \item \textbf{Cardboard crafts} (\textbf{\term{\cardboard}}): Crafting Omikuji boxes, Japanese random fortunes, from cardboard.
 \end{itemize}
These tasks satisfy the above criteria; the domains are diverse (electronics, chemistry, assembly, and crafts) and contain various mistakes (\eg, accidentally cutting off cardboard, using wrong chemicals). In addition, the workers should follow procedural texts to accomplish the task, and the video duration ranges from a few to 30 minutes.

\subsection{Preparation of procedural texts}
One of the authors wrote a procedural text for each task before recordings.
First, they prepare the draft versions referring to the following existing texts.
For color mixture experiments and cardboard crafts, they searched the web to collect procedural texts.
For ion reaction experiments and electrical circuits, they borrowed the texts from out-of-the-box kits.
For toy building blocks, they wrote procedures from scratch because we found no resources on the web or in kits.
We further revised these drafts to improve clarity (\eg specify the tools to be used).
These original procedural texts were written in Japanese, then we manually translated them to prepare the English versions.
The English version of procedural texts is bundled into the released dataset, and participants of our recording referred to the Japanese version as explained in the next section.

\subsection{Video recording}

\noindent\textbf{Participants, camera, and environments.}
Four Japanese graduate students (4 males) performed the tasks following procedural texts. The participants were equipped with a head-mounted camera Panasonic HX-A500 as shown in \cref{fig:camera}. It is a 30 fps video camera with 4K RGB resolution. 
When recording the tasks, the participants referred to the Japanese versions of procedural texts. We also gave them images of the finished products in electrical circuits and toy building blocks. This is because our preliminary experiments showed that the two tasks were difficult to complete only with the texts.
To avoid the influence of background changes, a desk with objects, tools, and printed procedural texts is placed in the same indoor position across the recording sessions. The participants are recorded sitting to capture manipulated objects in detail.

% \begin{figure}[t!b]
%     \centering
%     \includegraphics[width=0.8\columnwidth]{figures/camera-crop.pdf}
%     \caption{Egocentric head-mounted camera on participants.}
%     \label{fig:camera}
% \end{figure}

\begin{figure}[t!b]
    \centering
    \includegraphics[width=0.8\columnwidth]{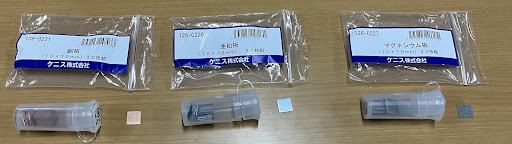}
    \caption{Raw visual cues of copper (left), zinc (center), and magnesium (right) plates. Especially, zinc and magnesium are indistinguishable.}
    \label{fig:copper-zinc-magnesium}
\end{figure}

\begin{figure}[t!b]
    \centering
    \includegraphics[width=0.8\columnwidth]{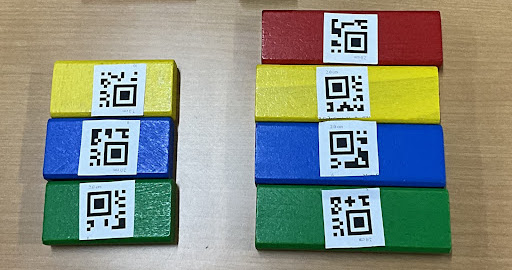}
    \caption{Micro QR codes on objects. It's difficult to specifically identify the short thin cuboids (left) and the long thin cuboids (right) only from vision. Micro QR codes are useful for distinguishing them.}
    \label{fig:qr}
\end{figure}

\noindent\textbf{Micro QR codes.}
Detecting objects mentioned in procedural texts from videos is crucial for identifying incorrect actions. For example, object detection can determine if a worker follows the instruction to paste papers using glue not tape. However, some objects are visually indistinguishable as shown in \cref{fig:copper-zinc-magnesium}.
Previous studies~\cite{naim_discriminative_2015,nishimura_biovl2_2022,schoonbeek_industreal_2024,lee_error_2024} assumed manual annotations of the objects, yet it's costly and time-consuming.
To address these issues, we attach Micro QR Codes~\cite{isoiec_180042024_information_2024} that encode the objects' names as shown in \cref{fig:qr}. This is an effortless solution because QR Code detection does not require further tuning for our dataset to apply existing detectors. Nevertheless, we cannot integrate QR Codes with our proposed approach now, thus our future work is to leverage QR Codes for object-oriented mistake detection.

\noindent\textbf{Recording process.}
A participant worked on every task 2 or 4 times, totaling 10 recordings for each task.
To record various mistake actions, they intentionally perform mistake actions that we assume in advance (\eg, skipping several procedures).
Out of ten times to record videos, they contained the intentional mistakes five times and tried to follow procedures correctly for the other five times. Note that accidental mistakes due to carelessness also happened even in the latter, which is desirable to contain natural mistakes in EgoOops.

\subsection{Annotation guidelines}

\noindent\textbf{Video-text alignment.}
We first annotate video-text alignment by extracting start and end timestamps (i.e., segments) and mapping them to the corresponding steps. To reduce the ambiguity of segment annotations, we define the segment period as a time range from grasping the step-related objects to releasing them. We also localize extra step segments not written in a procedural text, and they are labeled as \term{undefined} (\eg, grasping a wrong object, correcting a previous mistake).
Moreover, workers pause some steps and resume them after others (\term{split}), skip necessary steps (\term{missing}), and swap the order from the prescribed one (\term{out-of-order}). These are kinds of order mistakes, yet we do not attach their labels explicitly because the video-text alignment reveals them. For example, if step 2 occurs before step 1, these swapped steps are considered \term{out-of-order}.

\noindent\textbf{Mistake labels.}
If a segment indicates an execution mistake, we assign one label of the six execution mistake classes listed in \cref{subsec:supp-term}. This categorization can be used to analyze mistake patterns across diverse tasks (\eg, a commonly frequent category of mistakes). We observe multiple mistakes in just a few segments, so we label them as \term{\others} to avoid multi-label classification because the problem is difficult for current action understanding models.

\begin{figure*}[t!b]
    \centering
    \begin{subfigure}{0.33\linewidth}
        \centering
        \includegraphics[width=\linewidth]{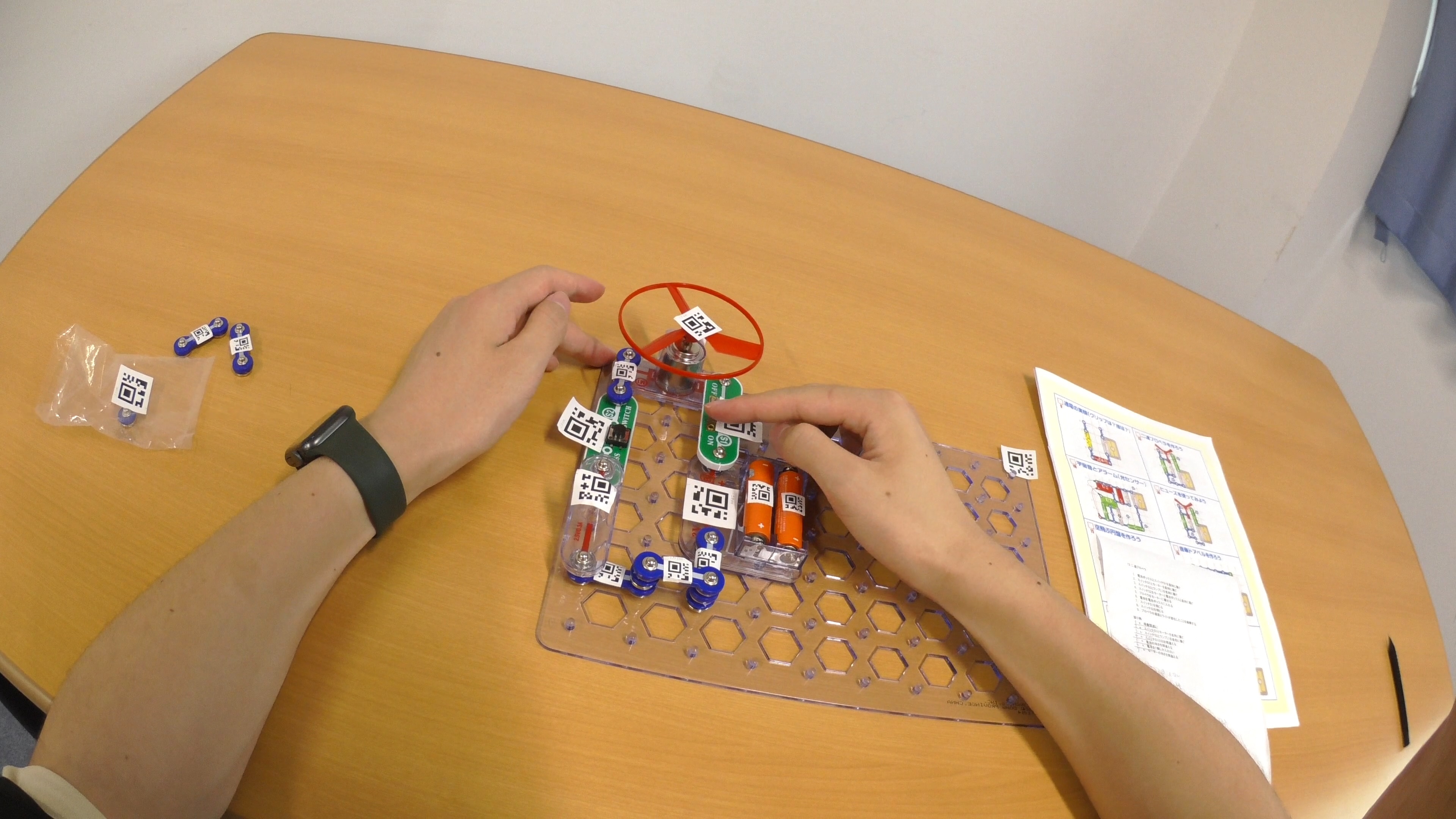}
        \caption{Described as ``put batteries in the wrong direction'' (mistake label: \term{\howto}).\\}
        \label{fig:desc-eg-battery}
    \end{subfigure}
    \hfill
    \begin{subfigure}{0.33\linewidth}
        \centering
        \includegraphics[width=\linewidth]{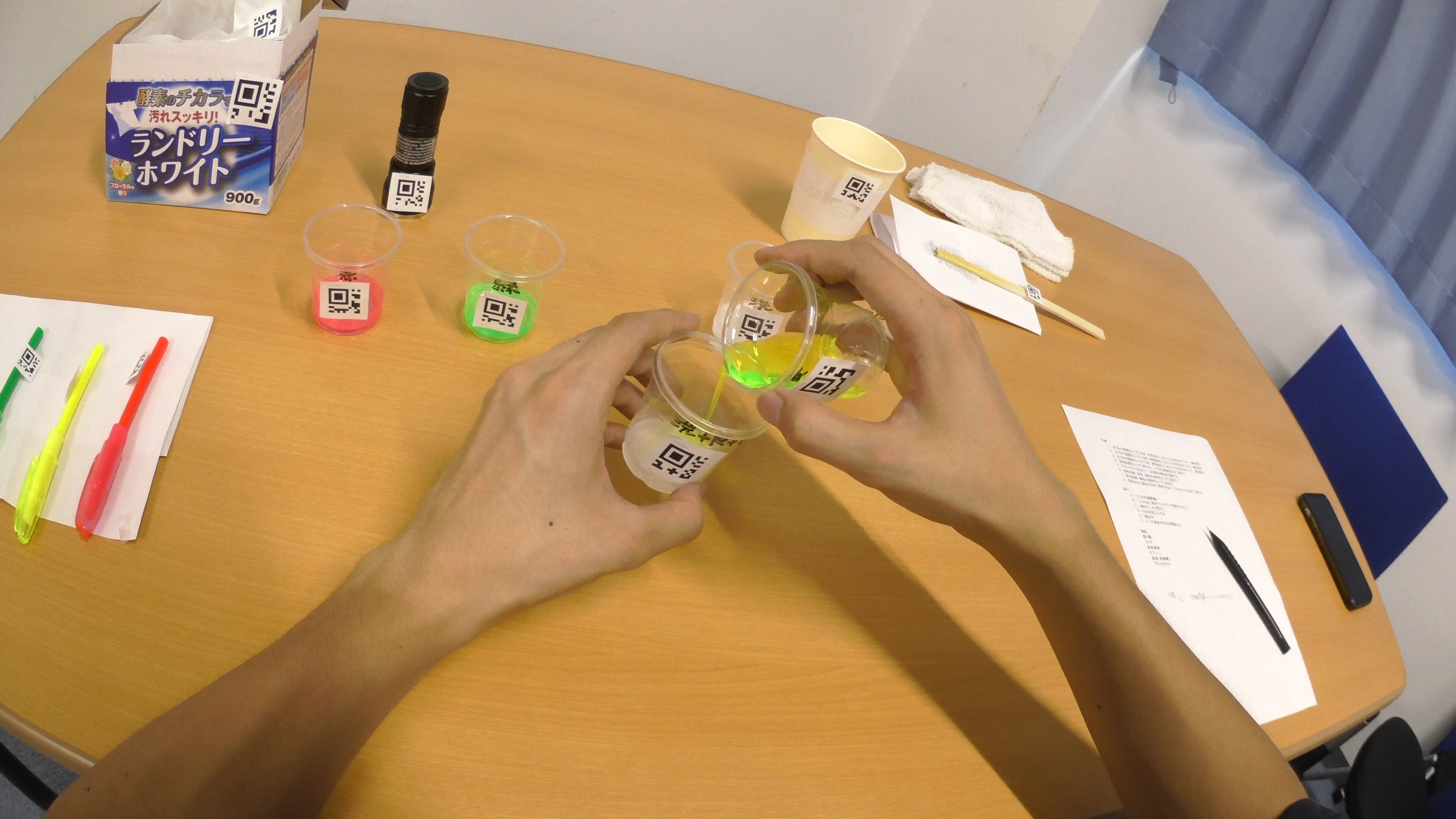}
        \caption{Described as ``use yellow liquid but should use green liquid'' (mistake label: \term{\object}).\\}
        \label{fig:desc-eg-liquid}
    \end{subfigure}
    \hfill
    \begin{subfigure}{0.33\linewidth}
        \centering
        \includegraphics[width=\linewidth]{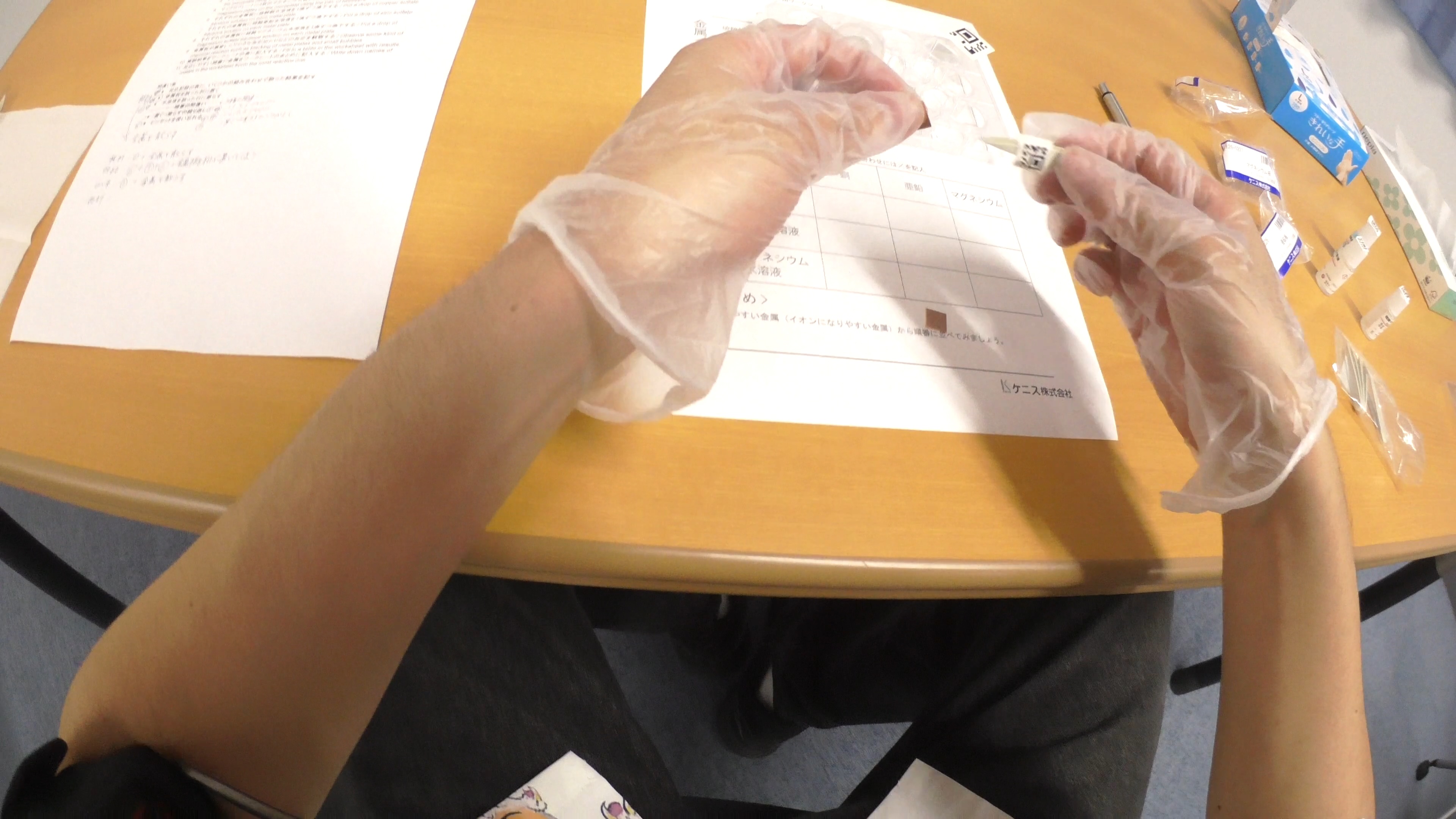}
        \caption{Described as ``drop copper plates on the table then pick them up with fingers'' (mistake label: \term{\accident}).}
        \label{fig:desc-eg-plate}
    \end{subfigure}
    \caption{Examples of mistake descriptions with video segments.}
    \label{fig:desc-eg-video}
\end{figure*}

\noindent\textbf{Descriptions.}
We also provide the mistake segments with descriptions of why the performed actions are considered mistakes. They can be used to analyze model performance in detail or interact with workers through smart assistants. To maintain the quality of the descriptions, a template for each mistake class minimizes the use of modifiers, subjects, and articles. \Cref{tab:desc_temp} shows description templates. The descriptions are written in English based on the templates, which are created based on mistake label classes. \Cref{fig:desc-eg-video} shows description examples annotated using the description templates.

\begin{table*}[t!b]
    \centering
    \caption{Description templates.}
    \label{tab:desc_temp}
    \begin{tabular}{l|c}
    \hline
    Mistake label & Description \\ \hline
    1. \object & Use \{wrong object\} but should use \{correct object\} \\
    2. \mispick &  Grasp \{wrong object\} \\
    3. \correction & Correct error in step \{step number\} \\
    4. \accident & \{verb\} (+\{object\}) \\
    5. \howto & \{verb\} \{object\} \{adverb\} but should \{verb\} it/them \{adverb\} \\
    6. \others & Free writing. Multiple sentences are allowed to describe multple mistakes. \\ \hline
    \end{tabular}
\end{table*}

\subsection{Annotation Tool}

\begin{figure*}[t!b]
    \centering
    \includegraphics[width=\linewidth]{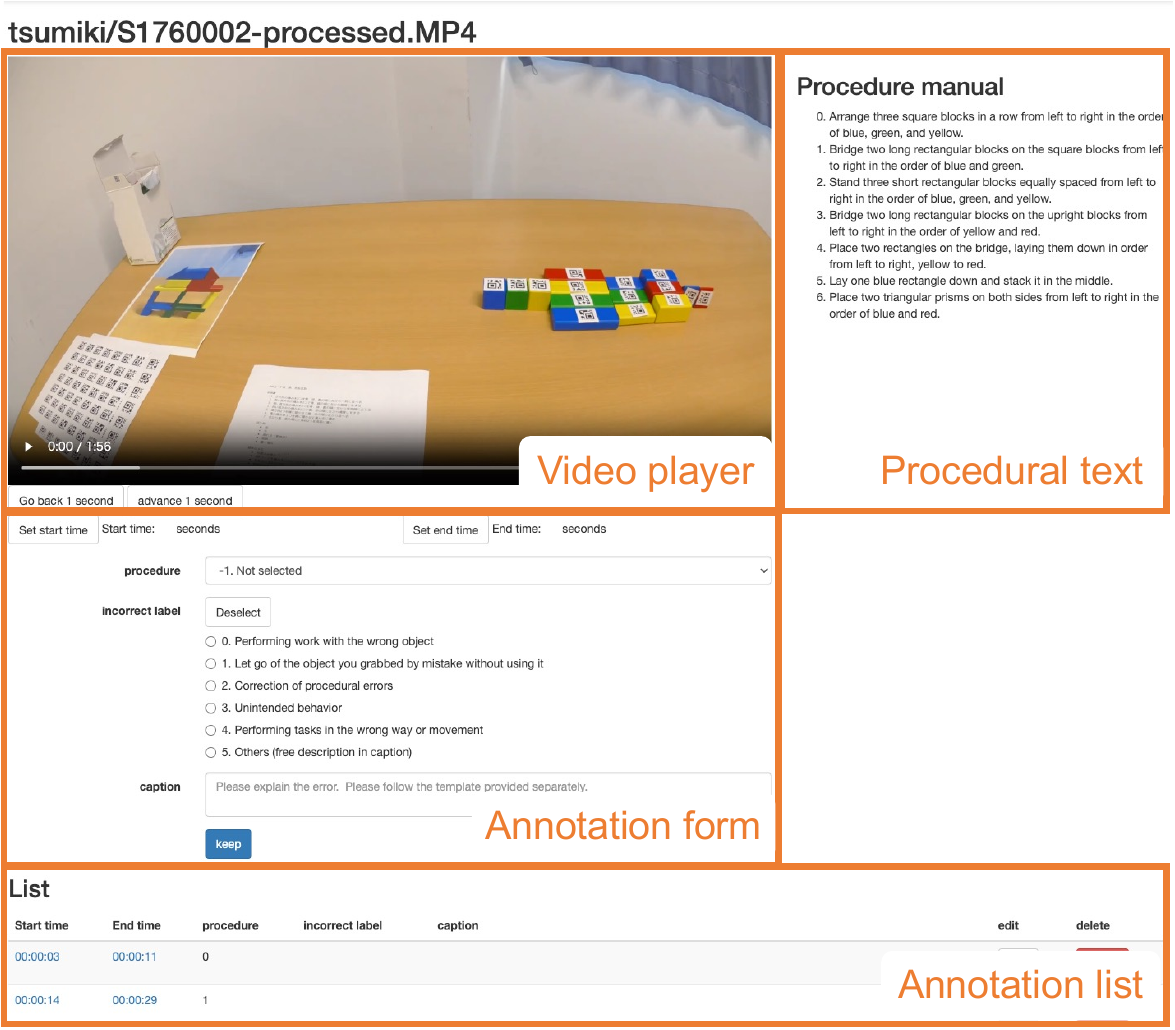}
    \caption{Overview of annotation tool. An annotator can watch a video (upper left) and read a procedural text (upper right). In the annotation form (middle), the annotator attaches video-text alignment, mistake labels, and descriptions. The annotations are stored in the annotation list (bottom).}
    \label{fig:annotation-tool}
\end{figure*}

\Cref{fig:annotation-tool} shows an annotation web tool, which consists of four panes: video player, procedural text, annotation form, and annotation list. In the video player and procedural text panes, annotators can watch the video and read the procedural text. For video-text alignment annotations, we implemented a feature that allows them to move one second backward or forward in the video. In the annotation form pane, the annotators attach video-text alignment, mistake labels, and descriptions. For video-text alignments, annotators select the steps corresponding to the video segments. For mistake labels, annotators choose from predefined labels. For descriptions, annotators write explanations detailing why the segments are mistakes. In the annotation list pane, the annotators can view the annotation results. They can also edit or delete them by clicking on the buttons.

\section{Dataset analysis}

\subsection{Inter-annotator agreement}

We ensure the quality of the original annotations by testing for agreement with another annotator. Since annotating all of the samples is time-consuming, we randomly choose one out of 10 videos per task. The process involves the following two types of annotations by the new annotator.
For video-text alignment, the annotator newly extracts segments and maps them to the corresponding steps. In addition, they watch the original segments and give new mistake labels and descriptions. To evaluate the annotation quality of video-text alignment, we calculate the temporal Intersection over the Union (tIoU) of segments. Note that we ignore the undefined steps for the calculation. We first calculate tIoU for each step, average by videos, and finally average them as an aggregated score.
As for mistake labels and descriptions, we calculate Cohen's kappa~\cite{cohen_coefficient_1960} and BERTScore~\cite{zhang_bertscore_2020}, respectively.

\noindent\textbf{Results.}
We confirm that both annotations are high quality. For the video-text alignment, we achieve 88.8 tIoU, and for mistake labels and descriptions, we record 86.8 Cohen's kappa and 96.3 BERTScore. These high scores ensures the high agreement between the original and new annotations.

\subsection{Statistics}

\subsubsection{Distribution of order mistakes}

% \Cref{tab:mistake-stats-task}に，各作業ごと及び全ての作業について，\cref{subsec:term}で述べた4種類の\ordermistakes 及びその合計の数を示している．全て合わせると，138回の\ordermistakes が生じている．誤りの種類ごとにみると，\term{\resume}が最も多く，合計で51回生じている．次に多く生じているのが，\term{\outoforder}であり，合計で40回である．作業ごとにみると，最も順序誤りが多いのは，段ボール工作の61回である．その中でも上位2種類の誤りは，\term{\resume}と\term{\outoforder}であり，作業者が手順書内の手順を自由な順番で実行していると読み取れる．対照的に，光の混色反応実験とイオン反応実験では，\term{\undefined}が最も多く生じており，余分な手順を挟んでしまっていることが分かる．このように，それぞれの作業は，\ordermistakes に関して，異なる傾向を示している．

% \Cref{tab:mistake-stats-task} shows the four types of \ordermistakes mentioned in \cref{subsec:term} and the total number of them for each task and for all tasks. In total, 138 \ordermistakes occurred. Looking at each type of error, \term{\resume} was the most common, occurring 51 times in total. The next most common was \term{\outoforder}, occurring 40 times in total. Looking at each task, the most common error was cardboard work, with 61 errors. The top two types of errors were \term{\resume} and \term{\outoforder}, which suggests that workers were performing the steps in the instructions in a free order. In contrast, \term{\undefined} was the most common error in the light mixing reaction experiment and the ion reaction experiment, which suggests that extra steps were inserted. As such, each task shows different trends in terms of \ordermistakes.

In \cref{tab:mistake-stats-task}, we provide the number of the four types of \ordermistakes listed in \cref{subsec:supp-term}. The total count of \ordermistakes is 138. Looking at each type of error, \term{\resume} is the most common type, occurring 51 times in total. The next most common one is \term{\outoforder}, occurring 40 times in total. Looking at each task, the task where the mistakes happen most frequently is the cardboard crafts, with 61 errors. The top two types of mistakes in the task are \term{\resume} and \term{\outoforder}, which suggests that workers perform the steps in the instructions in a free order. In contrast, \term{\undefined} is the most common mistake in the color mixture experiments and the ionic reaction experiments, which suggests that extra steps are inserted. As such, each task shows different trends in terms of \ordermistakes.

\begin{table}[t!b]
    \centering
    \caption{The number of \ordermistakes.}
    \label{tab:mistake-stats-task}
    \scalebox{0.78}{
    {\tabcolsep = 1mm
    \begin{tabular}{l|rrrr|r} \toprule
        \multirow{2}{*}{Task} & \multicolumn{5}{c}{\ordermistakes}\\
         & \missing & \outoforder & \resume & \undefined & Total \\ \midrule
        \electricalcircuit & 2 & 10 & 12 &  6 & 30 \\
        \colormixture      & 1 &  2 &  3 &  9 & 15 \\
        \ionicreaction     & 2 &  4 &  1 &  6 & 13 \\
        \buildingblock     & 0 &  3 &  6 & 10 & 19 \\
        \cardboard         & 7 & 21 & 29 &  4 & 61 \\ 
        \midrule
        All                &12 & 40 & 51 & 35 &138 \\
        \bottomrule
    \end{tabular}
    }
    }
\end{table}

\subsubsection{Verb and noun distributions of mistake descriptions}

\begin{figure*}[t!b]
    \centering
    \begin{subfigure}{0.40\linewidth}
        \centering
        \includegraphics[width=\linewidth]{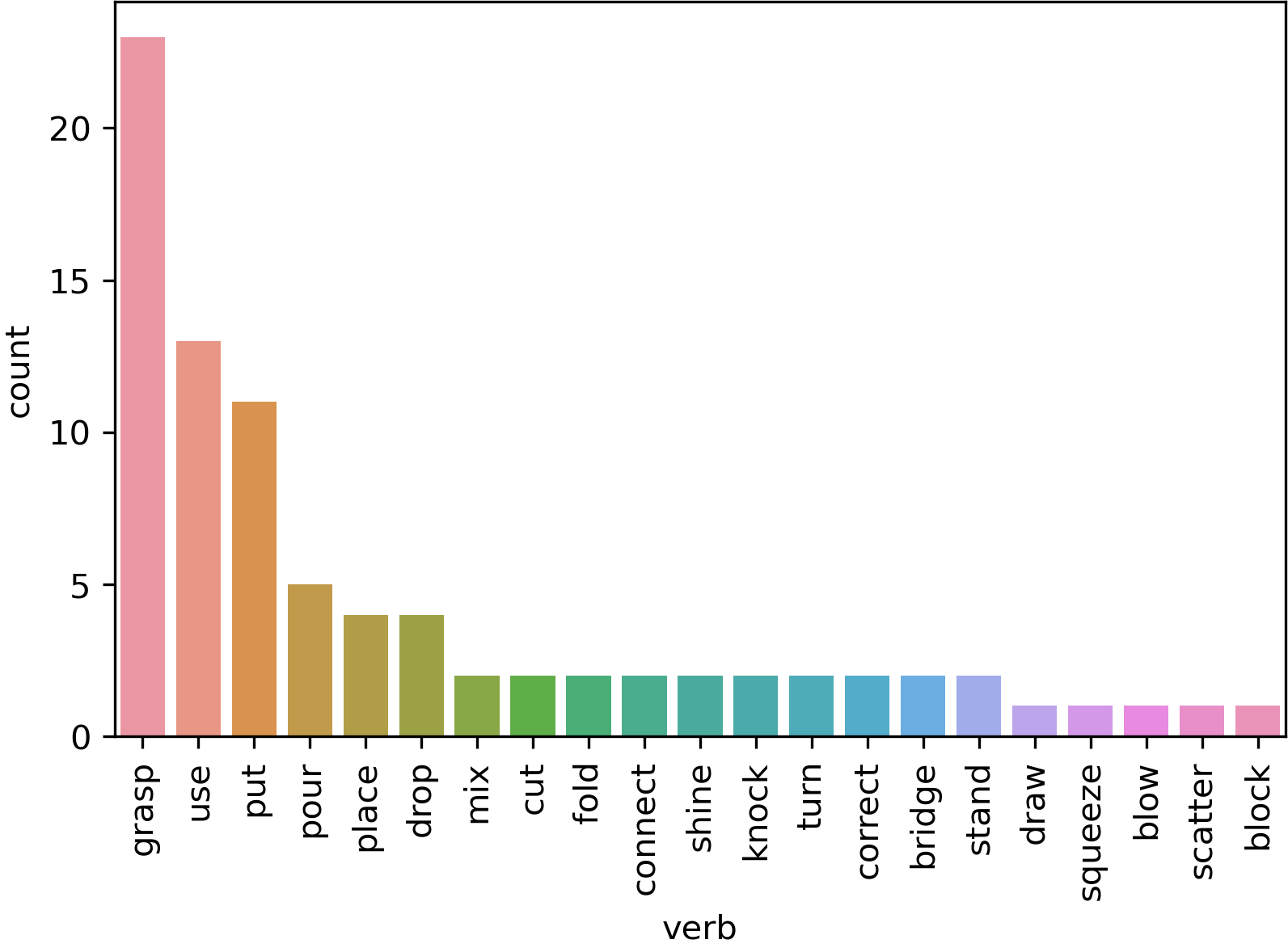}
        \caption{Distribution of verbs.}
        \label{fig:desc-dist-verb}
    \end{subfigure}
    \hspace{1mm}
    \begin{subfigure}{0.58\linewidth}
        \centering
        \includegraphics[width=\linewidth]{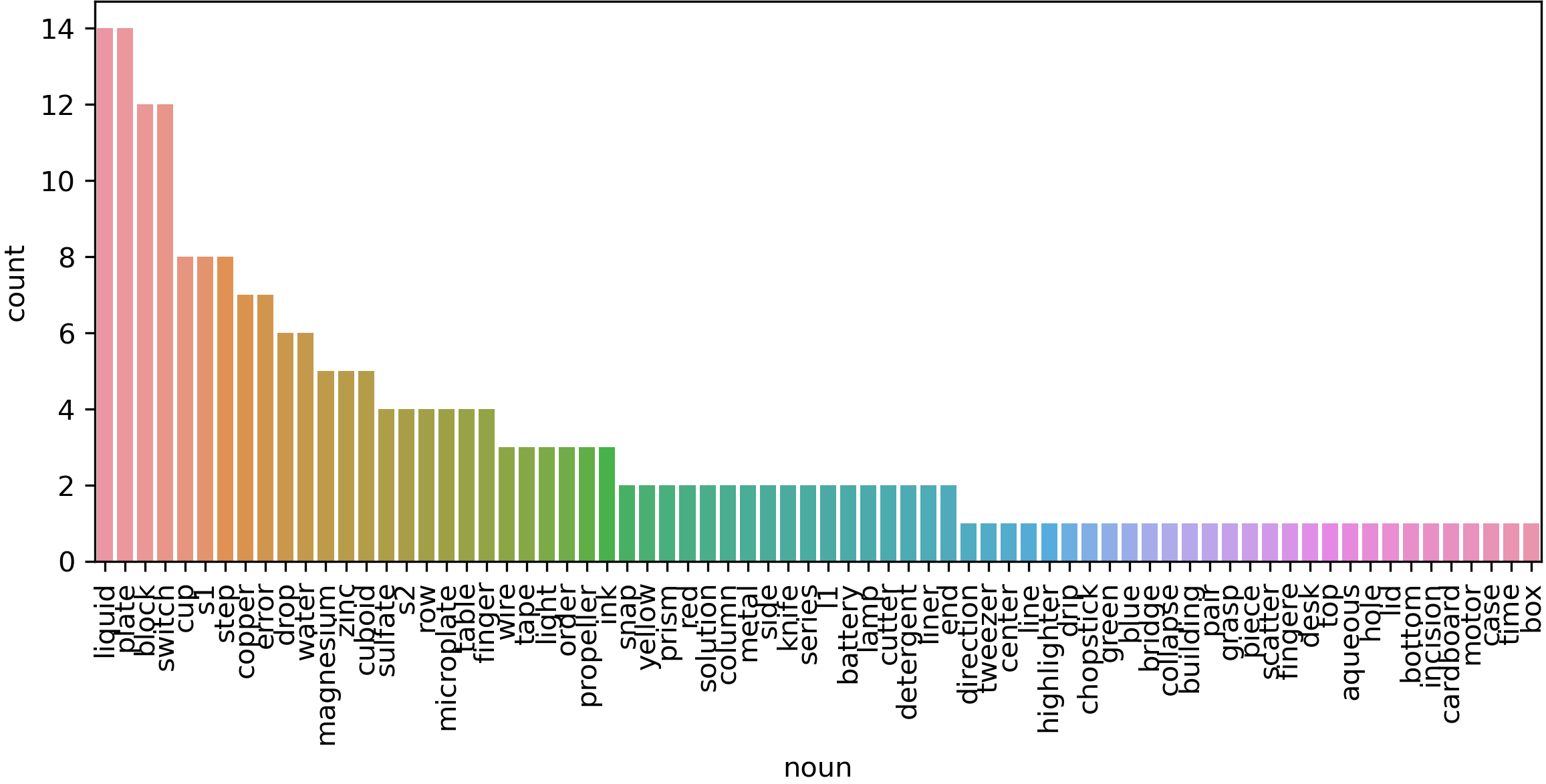}
        \caption{Distribution of nouns.}
        \label{fig:desc-dist-noun}
    \end{subfigure}
    \caption{Distribution of verbs and nouns in mistake descriptions.}
    \label{fig:desc-dist-word}
\end{figure*}

\Cref{fig:desc-dist-word} shows the verb/nouns distributions of descriptions. We extract verbs and nouns using spaCy\footnote{\url{https://spacy.io/}} based on \texttt{en\_core\_web\_trf}.
Note that for descriptions with the mistake labeled as \term{1. object} or \term{5. way}, we ignore the phrase after ``but'' because this phrase is not a mistake but rather a description of the correct step.

Regarding verbs, the top two verbs, ``grasp'' and ``use,'' are the words in the templates. The third most frequent verb is ``put,'' indicating mistakes related to the location and direction of objects (\eg, ``put batteries in the wrong direction'' in the electrical circuits). This description can be confirmed in \cref{fig:desc-eg-battery}.
Regarding nouns, ``liquid'' and ``plate'' are the most frequent.
These mistakes are common in color mixtures and ionic reactions (\eg, ``use yellow liquid but should use green liquid'' in the color mixture in \cref{fig:desc-eg-liquid}.
``Block'' and ``switch'' are two nouns of the third most frequent, representing mistakes in building blocks and electrical circuits, respectively.

\section{Dataset publication}

We have published EgoOops to accelerate future studies for mistake action detection and reduce heavy burdens on workers due to mistakes. It can be obtained without a personal request in the expectation of effortless access. 

\subsection{Completeness of the relevant documentation}

\Cref{sec:supp-dataset-construction} describes how the data was collected and organized and what information it contains. We recommend users refer to the README document bundled with the published annotations, which explains how the dataset can be read.

\subsection{Licensing and access}

% Hosting
We regard accessibility and long-term preservation as important. We provide access to the dataset through the project page\footnote{\url{https://y-haneji.github.io/EgoOops-project-page/}} hosted on the GitHub Pages\footnote{\url{https://pages.github.com}}. We host the annotations on a GitHub repository\footnote{\url{https://github.com/Y-Haneji/EgoOops-annotations/}}{\renewcommand{\thefootnote}{\fnsymbol{footnote}}\footnotemark[1]}{\renewcommand{\thefootnote}{\fnsymbol{footnote}}\footnotetext[1]{Refer to the project page for the latest URLs.}} because it presents an easy interface to update the data and show the history. We decided to host the videos on our laboratory's website\footnote{\url{https://www.lsta.media.kyoto-u.ac.jp/resource/data/EgoOops/}}{\renewcommand{\thefootnote}{\fnsymbol{footnote}}\footnotemark[1]} due to a file size limitation of Github.
% Maintenance
We will provide these data with the necessary maintenance following our organization's policy~\cite{kyoto_university_research_information_management_committee_policy_2020}.

% Licensing
EgoOops is licensed under CC BY-SA 4.0\footnote{\url{https://creativecommons.org/licenses/by-sa/4.0/}}. The license is a relatively permissive choice to accept wide uses because our intended use cases include commercial and industrial situations.

\subsection{Consent and privacy}

We asked the participants and obtained explicit consent to release the dataset that they had captured. Moreover, we managed to minimize the exposure of any personally identifiable information. Sounds are recorded but muted upon publication because workers sometimes spoke or talked to others. We trimmed unrelated segments to the task execution in the beginning and end of the videos (\eg, other workers' help with task preparation and camera operation). 

\subsection{Ethics and responsible use}

Our dataset is expected to be used for research on mistake detection. However, we do not prohibit the use of our dataset for other purposes as long as they are ethically acceptable. The entire part of our dataset was created for this study. Hence, it does not confront any ethical issues raised by other publicly released datasets.

\subsection{Legal compliance}

We ensure our awareness and compliance with regional legal requirements.

\subsection{Author statement on responsibility}

We state that we bear all responsibility in case of violation of rights, etc., and confirmation of the data license.

\section{Details of experiments.}

\subsection{Mistake label classification}

\noindent\textbf{Existing mistake classifiers.}
We test existing mistake classifiers proposed by and trained on Assembly101~\cite{sener_assembly101_2022}  and CaptainCook4D~\cite{peddi_captaincook4d_2024}. 
Sener \etal~\cite{sener_assembly101_2022} proposed to apply a long-range video model TempAgg~\cite{sener_temporal_2020} using TSM~\cite{lin_tsm_2019} features. TSM is a video recognition model and they trained it for action recognition on Assembly101. The features are extracted using the completed TSM and passed to TempAgg for mistake classification into three classes: \term{correct}, \term{mistake}, and \term{correction}. We had not found their official implementation for feature extraction and mistake classification and thus, decided to reproduce them by ourselves.\footnote{We used the TSM model weights provided by Sener \etal at \url{https://drive.google.com/file/d/11uFkqg1tWfWeATukjFL_HXHB6Gs1iGfU/view?usp=sharing}.}
As for a CaptainCook4D model, we chose the most performant variant of their error recognition models: a multi-layer perceptron head with a frozen 3D-ResNet~\cite{hara_learning_2017} backbone for feature extraction. We followed their official implementations for feature extraction\footnote{\url{https://github.com/CaptainCook4D/feature_extractors/}} and error recognition\footnote{\url{https://github.com/CaptainCook4D/error_recognition/}}.

\noindent\textbf{Multi-modal large language models (MLLMs).}
We evaluate two leading open-source MLLMs on EgoOops in a zero-shot manner aiming to inspect their visual reasoning capabilities; the two models are InternVL2.5-8B~\cite{chen_expanding_2024} and Qwen2-VL-7B-Instruct~\cite{wang_qwen2-vl_2024}. In our experiments, they solve mistake label classification given a trimmed video clip, the task's procedure, and the performed step. We construct the prompt by replacing placeholders in the template shown in \cref{fig:mllm-prompt-template}. We designed this template with the following aim. 
\begin{enumerate*}
    \item It contains the whole steps of the task to provide the context of the ongoing work.
    \item It encourages them to discover mistake actions and their corrections actively.
    \item It adopts multiple-choice questions as the direction format because MLLMs are trained to answer them well.
\end{enumerate*}
The completed prompt and the frames sampled from the video clip are passed to the models as inputs, and the models output their answers in free-style texts.
We used LMDeploy\footnote{\url{https://lmdeploy.readthedocs.io/en/latest/index.html}} for the inference processes to keep the experimental settings as fair as possible. For the pre-processing of the videos, we used their official implementations.\footnote{InternVL2.5: \url{https://github.com/OpenGVLab/InternVL/}}\footnote{Qwen2-VL: \url{https://github.com/QwenLM/Qwen2-VL/}} Specifically, for InternVL2.5, 24 frames are uniformly sampled from a video clip, and each of them is represented by 256 visual tokens; for Qwen2-VL, we sample each video at 2fps limiting the maximum number of frames to 48, and every frame is encoded to 128 visual tokens.

% \begin{figure*}[t!b]
% \centering
% \begin{tcolorbox}
% Procedure:\\
% \verb|{PROCEDURE}|\\
% This step: \verb|{STEP_INSTRUCTION}|\\
% It is an egocentric video clip where the worker performs an activity referring to the procedure. Note that if the step is "UNDEFINED", it is an extra step not written in the procedure.\\
% Carefully look at the clip. Try to find worker's failures of precisely carrying out the step instruction (i.e. mistake) or correction of mistakes. We penalize more for overlooking mistake and correction classes. Select the best option to the following multiple-choice question based on the video clip.\\
% Question: Which label best matches the activity performed by the camera wearer?\\
% 0. correct\\
% 1. correction\\
% 2. mistake\\
% The best option:\\
% \end{tcolorbox}
% \caption{The prompt template for MLLMs. We replace the placeholders with the task's procedure and the performed step for each trimmed video clip.}
% \label{fig:mllm-prompt-template}
% \end{figure*}

\noindent\textbf{Evaluation metrics.}
We report standard metrics for mistake classification problems~\cite{sener_assembly101_2022,wang_holoassist_2023,peddi_captaincook4d_2024,flaborea_prego_2024,seminara_differentiable_2024-1}: recall, precision, and F1 score. Since we aim to find mistake actions and their corrections, each metric is calculated only for the \term{mistake} and \term{correction} classes. Note that CaptainCook4D's model classifies each step into two classes: \term{normal} (\ie, \term{correct}) and \term{error} (\ie, \term{mistake})~\cite{peddi_captaincook4d_2024}, thus its evaluation results for the \term{correct} class are not available in our experiments. Also, we evaluate the predictions of MLLMs by checking whether the id and name of classes (\eg, \term{0. correct}) exactly match the ground truths.

\end{document}